\documentclass[11pt]{article}
\usepackage[letterpaper,margin=1in]{geometry}
\usepackage[T1]{fontenc}
\usepackage[utf8]{inputenc}
\usepackage{amsmath}
\usepackage{amssymb}
\usepackage{amsthm}
\usepackage{booktabs}
\usepackage{array}
\usepackage{tabularx}
\usepackage{multirow}
\usepackage{float}
\usepackage{graphicx}
\usepackage{caption}
\usepackage{subcaption}
\usepackage{pdflscape}
\usepackage{siunitx}
\usepackage{enumitem}
\usepackage{microtype}
\usepackage{xcolor}
\usepackage{url}
\usepackage{authblk}
\usepackage{natbib}

\bibpunct{(}{)}{;}{a}{}{,}
\setcitestyle{authoryear}

\usepackage{setspace}
\usepackage{hyperref}
\hypersetup{
    colorlinks=true,
    linkcolor=blue!70!black,
    citecolor=green!50!black,
    urlcolor=blue!70!black,
    breaklinks=true,
    bookmarksnumbered=true,
    pdftitle={Drive, Pack, Fly: The Travelling Thief Problem with Drone},
    pdfauthor={Kabir Murjani, Abhay Sobhanan}
}

\newif\ifmarkchanges
\markchangesfalse

\newcolumntype{C}[1]{>{\centering\arraybackslash}m{#1}}
\newcommand{\tabfit}{\ifdim\width>\textwidth\textwidth\else\width\fi}

\newcommand{\depot}{0}
\newcommand{\depotend}{0'}
\newcommand{\R}{\mathbb{R}}
\newcommand{\N}{\mathbb{N}}
\newcommand{\given}{\,|\,}

\theoremstyle{plain}
\newtheorem{proposition}{Proposition}
\theoremstyle{definition}

\title{\vspace{-1.5em}Drive, Pack, Fly: The Travelling Thief Problem with Drone}

\author[1]{Kabir Murjani}
\author[$*$\,2]{Abhay Sobhanan}
\affil[1]{Nirma University, SG Highway, Ahmedabad 382481, Gujarat, India}
\affil[2]{Indian Institute of Management Bangalore, Bannerghatta Road, Bengaluru 560076, Karnataka, India}

\date{\today}

\begin{document}

\maketitle
\thispagestyle{empty}
\renewcommand{\thefootnote}{\fnsymbol{footnote}}
\footnotetext[1]{Corresponding author. Email: \texttt{abhay.sobhanan@iimb.ac.in}}
\renewcommand{\thefootnote}{\arabic{footnote}}\setcounter{footnote}{0}

\begin{center}
\begin{minipage}{0.88\textwidth}
\begin{center}\textbf{Abstract}\end{center}
\noindent
In collection operations, accumulating payload progressively slows the vehicle, imposing a cumulative penalty on routing efficiency. An onboard drone can offset this penalty by retrieving outlying items, thereby shortening the makespan and increasing operational profit. However, travel time remains load-dependent, and each item collected by the ground vehicle shifts the arrival times that govern the drone's launch and rendezvous points. This paper introduces the Travelling Thief Problem with Drone (TTP-D), which maximises the collected profit, net of a time-based rental cost, by jointly optimising item selection, vehicle routing, and flight synchronisation. We formulate a mixed-integer linear program that solves small instances to optimality, and develop both metaheuristics and an attention-based Deep Reinforcement Learning (DRL) policy for larger instances. We further propose a learner-initialised hybrid solver, in which the DRL policy constructs an initial solution that a short annealing run subsequently refines. On two benchmark sets, this hybrid recovers most of the metaheuristic baseline's quality at a fraction of its computational budget, although the largest instances still require the baseline at its full budget. Finally, a sensitivity analysis reveals that the rental ratio is the primary driver of profitability, whereas the fleet parameters affect profit only at the margin. 
\end{minipage}
\end{center}

\vspace{1em}

\vspace{1em}

\section{Introduction}
\label{sec:introduction}

Drone logistics has transitioned from conceptual announcements to active field operations within a decade. While Amazon’s Prime Air established delivery drones in the public consciousness, autonomous aircraft now transport blood and medical supplies to rural clinics \citep{ackerman2019blood}. Additionally, truck-launched drones have been trialled by parcel carriers and extensively studied for relief distribution across damaged road networks \citep{otto2018optimization, chung2020optimization, sobhanan2024branch}. Since a drone carries a single light payload over a limited range, practical designs pair it with a road vehicle that hauls the heavy load and ferries the aircraft between launch points \citep{murray2015flying}. The truck handles the heavy cumulative load, while the drone reaches distant locations within range.

While these deployments focus on outbound delivery, this paper is motivated by the reverse operation: collection. Scenarios such as a courier picking up samples from rural clinics, a recycling vehicle emptying collection points, or a relief vehicle retrieving supplies across a damaged road network all share two critical features that are often omitted in classical routing models. First, the vehicle fills as it progresses, and this accumulating load incurs operational penalties. Fuel consumption rises, and travel speeds drop with added weight, meaning an early, heavy pickup taxes every subsequent leg of the tour. Second, an onboard drone can efficiently fetch light, high-value items from locations that are expensive to reach by road, but only if its flights are strictly synchronised with the truck's itinerary. Since fleet deployment incurs operational costs, time savings anywhere within the schedule translate directly into financial value.

Existing literature addresses these two features in isolation. The Travelling Thief Problem (TTP), introduced by \citet{bonyadi2013travelling}, couples the Travelling Salesman Problem (TSP) with a Knapsack Problem (KP) through a weight-dependent velocity function: a vehicle's speed decreases as its load increases, and only a subset of items is collected, subject to capacity constraints. However, the model omits multi-vehicle dynamics. Conversely, truck-and-drone routing captures synchronised sorties but is generally designed solely for delivery \citep{mahmoudinazlou2024hybrid}. This paper introduces the Travelling Thief Problem with Drone (TTP-D), bridging the gap to address this combined collection and routing setting that requires truck-drone coordination.

In the TTP-D, a capacitated truck and a single-package drone are based at a common depot. All $N$ customers are visited, one vehicle travelling to each, but the fleet collects the item at only a subset of them. The drone can be launched from a node the truck has visited to pick up an item at an outlying customer location, then hand it over to the truck at a subsequent rendezvous node farther along the route. The operator must decide which customers each vehicle visits, which items to collect subject to the truck's capacity and the drone's payload limits, and where the two vehicles coordinate launches and rendezvous. The objective is to maximise the collected profit minus a routing cost proportional to the makespan, the time at which the fleet completes the mission. The problem is $\mathcal{NP}$-hard, as it generalises both the TSP and the 0-1 KP. These subproblems are tightly coupled: collecting an additional item shifts all subsequent arrival times and can invalidate a drone rendezvous planned further downstream. This coupling of the decision space constitutes a fundamental source of difficulty
in the TTP-D.

We formulate the TTP-D as an exact mixed-integer linear program (MILP) that provides optimality certificates for small instances. To overcome the model's computational intractability on larger benchmarks, we design three scalable algorithmic approaches. First, we develop Simulated Annealing (SA) and Variable Neighbourhood Search (VNS) metaheuristics, which trade optimality for computational efficiency. Second, we employ Deep Reinforcement Learning (DRL) to learn construction policies \citep{kool2019attention}, amortising the search cost into an offline training phase and yielding solutions for a fixed, minimal decoding budget. Finally, we introduce a hybrid framework that bridges learning and search. A DRL policy trained via behaviour cloning on metaheuristic solutions generates a high-quality initial plan, which is subsequently refined by a brief local search. On our benchmarks, this hybrid recovers the performance of the baseline metaheuristic to within mean gaps of $1.6\%$ and $5.1\%$ using half of its computational budget. Throughout our computational study, the exact solver anchors the comparison using provable bounds where available, while the scalable methods are evaluated by their performance gaps to the best-known solutions.

This study makes the following contributions.
\begin{itemize}
  \item We introduce the Travelling Thief Problem with Drone (TTP-D), coupling the load-dependent routing and packing dynamics of the TTP with time-synchronised drone sorties. We formulate the problem as an MILP featuring a piecewise-linear velocity approximation.
  
  \item We develop two scalable and effective metaheuristics, Simulated Annealing and Variable Neighbourhood Search, equipped with a common neighbourhood library and an efficient, feasibility-preserving evaluation scheme. 
  
  \item We cast the TTP-D as a Markov Decision Process (MDP) and train a DRL construction policy. We compare a graph attention (GAT) encoder against a multilayer perceptron (MLP) encoder, training both via Proximal Policy Optimisation (PPO) with a Policy Optimisation with Multiple Optima (POMO) baseline.
  
  \item We introduce LISA (Learner-Initialised Simulated Annealing), a hybrid algorithm that distils the metaheuristic into a neural policy via behaviour cloning and couples the two during inference. A single budget parameter dictates the balance between the neural policy and the search, while achieving performance comparable to a standalone annealing run with only 5–50\% of the runtime budget.
  
  \item We present a comprehensive computational study across two benchmark families: the well-known \texttt{a280}-derived sets, and \texttt{ttd300}, a newly simulated suite of drone endurance-controlled instances. Furthermore, our sensitivity analysis demonstrates that the renting ratio determines overall profitability and that drone endurance limits the value of higher speeds.
\end{itemize}

The remainder of the paper is organised as follows. Section~\ref{sec:related} reviews the related literature. Section~\ref{sec:problem} describes the problem statement and its assumptions, while Section~\ref{sec:MILP} presents the mathematical formulation. Section~\ref{sec:methodology} develops the metaheuristics, DRL framework, and the LISA hybrid solver. Sections~\ref{sec:results} and~\ref{sec:sensitivity} report the computational study and the sensitivity analysis, respectively, with detailed results and additional ablations provided in the Supplementary Material. Finally, Section~\ref{sec:conclusions} concludes the paper.

\section{Literature Review}
\label{sec:related}

We review the literature relevant to the TTP-D and our proposed algorithm across four research streams. Section~\ref{sec:related-ttp} reviews the TTP, the single-vehicle variant of our problem.  Section~\ref{sec:related-drone} covers truck-and-drone collaborative routing, forming the basis of our synchronised sortie operations. Finally, Sections~\ref{sec:related-learning} and~\ref{sec:related-hybrid} examine learning-based methods and hybrid algorithms that motivate the design of our construction policy and the LISA solver.

\subsection{The Travelling Thief Problem}
\label{sec:related-ttp}
The TTP \citep{bonyadi2013travelling} captures the interdependence between coupled optimisation subproblems by combining the TSP with the Knapsack Problem \citep{martello1990knapsack}. A thief visits a set of cities, selecting some items along the way, each of which increases the knapsack weight. A weight-dependent velocity function consequently slows the thief as the knapsack fills. Routing and packing decisions are therefore inseparable, as the value of a packing plan depends on the route that carries it, and the cost of a route depends on the items selected. \citet{polyakovskiy2014comprehensive} introduced a benchmark suite of 9,720 instances by assigning items to cities from the TSP Library (TSPLIB) \citep{reinelt1991tsplib}. This suite became the standard benchmark, from which the \texttt{a280} instance used in this study is drawn. Multiple heuristic methods subsequently emerged, including approximate constructive approaches \citep{faulkner2015approximate}, local-search and simulated-annealing hybrids \citep{elyafrani2018efficiently}, and per-instance selection portfolios \citep{wagner2018case}. However, these methods rely on hand-crafted rules, whereas exact mixed-integer formulations remain intractable beyond small instances.

\subsection{Truck-and-Drone Collaborative Routing}
\label{sec:related-drone}

\citet{murray2015flying} pioneered truck-and-drone collaborative routing with the Flying Sidekick TSP, in which a truck and drone serve customers in parallel. This framework was subsequently expanded: \citet{agatz2018optimization} applied exact and dynamic programming methods to the TSP with Drone. Driven by significant efficiency gains, research now encompasses diverse settings, including multi-drone fleets \citep{murray2020multiple, nguyen2022min}, two-echelon systems \citep{zhou2023exact}, and arc routing \citep{sobhanan2025arc}. Comprehensive surveys by \citet{otto2018optimization} and \citet{chung2020optimization} map the breadth of this literature and its various applications. The core computational challenge is spatio-temporal synchronisation: although traversing distinct routes, the vehicles must coordinate at rendezvous nodes, tightly coupling their routing decisions. In contrast to our work, this literature predominantly focuses on pure delivery routing problems without knapsack considerations. 

\subsection{Learning-Based Solvers}
\label{sec:related-learning}
Neural combinatorial optimisation has emerged as a compelling alternative to hand-crafted rules, utilising learned policies to construct solutions directly. Pointer Networks \citep{vinyals2015pointer} first demonstrated that attention mechanisms could output input permutations. \citet{bello2016neural} subsequently trained these networks via policy-gradient reinforcement learning for the TSP and knapsack problems. 
Later, the Attention Model \citep{kool2019attention} replaced recurrence with a Transformer encoder, achieving near-optimal tours across various routing variants. In contexts closer to the TTP-D, \citet{santiyuda2024solving} applied multi-objective reinforcement learning to the bi-objective TTP.
Traditional encoders lack the capacity to natively output coupled tour and packing sequences; consequently, the authors adopted an encoding-decoding scheme and trained only on small instances. Similarly, \citet{bogyrbayeva2023deep} generated near-instantaneous, competitive solutions for the TSP with Drone using an attention encoder paired with a Long Short-Term Memory (LSTM) decoder. This learning-based construction paradigm is also effective in adjacent dynamic logistics domains, such as order picker routing in warehouses \citep{mahmoudinazlou2025deep}.

\subsection{\texorpdfstring{Neural Hybrid Solvers}{Neural Hybrid Solvers}}
\label{sec:related-hybrid}
Bridging hand-built search and end-to-end learning is a family of hybrid approaches \citep{bengio2021machine}: learning branching decisions inside exact search \citep{khalil2016learning}, selecting the repair operator of a large-neighbourhood search \citep{hottung2020neural}, and searching on top of a trained construction policy via sampling, beam search, or test-time adaptation of embeddings \citep{joshi2019efficient, choo2022simulation, hottung2022eas}.
Imitation learning offers a complementary route into a construction policy. For instance, \citet{joshi2019efficient} fit a graph network to optimal tours, \citet{fu2021generalize} combine a small supervised model with Monte Carlo tree search to reach far larger instances, and NeuroLKH \citep{xin2021neurolkh} learns the edge candidates that steer the Lin-Kernighan-Helsgaun heuristic; portfolio designs that run several solvers and retain the best answer have precedent on the TTP itself \citep{wagner2018case}.
LISA, our proposed hybrid solver, arranges these roles differently: the metaheuristic serves as the trainer, whose solutions are cloned into a construction policy via imitation. At solve time, a fraction of the metaheuristic’s runtime budget is allocated to repairing the cloned plan, enabling performance comparable to a full SA run under the same time limit.

\section{Problem Description}
\label{sec:problem}

We consider a truck-drone collaborative routing and collection problem involving a single truck with a knapsack capacity $W$ and a single drone with a payload limit $W^{\mathrm{D}}$. Both vehicles start from an origin depot $\depot$ to visit a set of customers. The drone can be dispatched from the depot or from any node at which the truck is present; it flies to a target customer, retrieves an item, and rejoins the truck at a subsequent rendezvous node along the truck's route. The fleet must ultimately return to the depot. The truck's velocity decreases linearly with its accumulated load, whereas the drone is assumed to maintain a constant velocity. The objective is to maximise net profit, defined as the total collected profit minus a rental cost proportional to the makespan:
\[
G \;=\; \sum_{i \in \mathcal{N}} p_i z_i \;-\; R \cdot \tau_{\depotend}
\]
where $p_i$ is the profit of item $i$, $z_i \in \{0,1\}$ is a binary variable indicating whether item $i$ is collected, $\tau_{\depotend}$ is the arrival time at the return depot $\depotend$, and $R$ is the renting ratio. Note that the origin depot $\depot$ and destination depot $\depotend$ serve as the source and sink nodes of the same physical location, respectively, both having zero profit and zero weight.

Throughout the paper, we distinguish between visitation decisions and collection decisions. A customer is \emph{visited} when a vehicle travels to its location. Every customer is visited exactly once, either by the truck or by the drone, and the assigned vehicle determines the customer’s \emph{visit mode}. The item at a visited customer is \emph{collected} only if the packing decision selects it. Hence, visiting a customer does not imply collecting their items; only a subset of all the items is collected. 

As noted in Section~\ref{sec:introduction}, the TTP-D is an $\mathcal{NP}$-hard problem featuring strongly interdependent routing and packing decisions. Non-linear travel times due to payload-dependent truck speeds complicate truck-drone synchronisation. Consequently, the model must simultaneously optimise customer-visit modes (truck, drone, or rendezvous), packing configurations, and vehicle routes, as well as temporally valid launch-and-rejoin pairings.

\begin{figure}[htbp]
\centering
\includegraphics[width=0.5\textwidth]{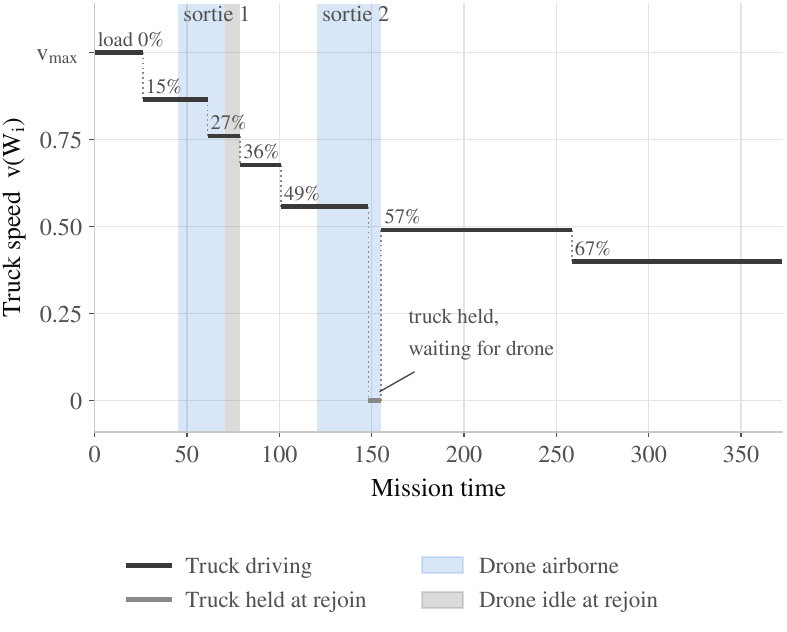}
\caption{Truck speed over a ten-customer TTP-D instance, drawn against mission
  time, with the two sortie windows shaded, waiting shown in grey, and the
  truck's load annotated as a percentage of capacity. }
\label{fig:ttpd_coupling}
\end{figure}

\begin{figure}[htbp]
\centering
\includegraphics[width=0.7\textwidth]{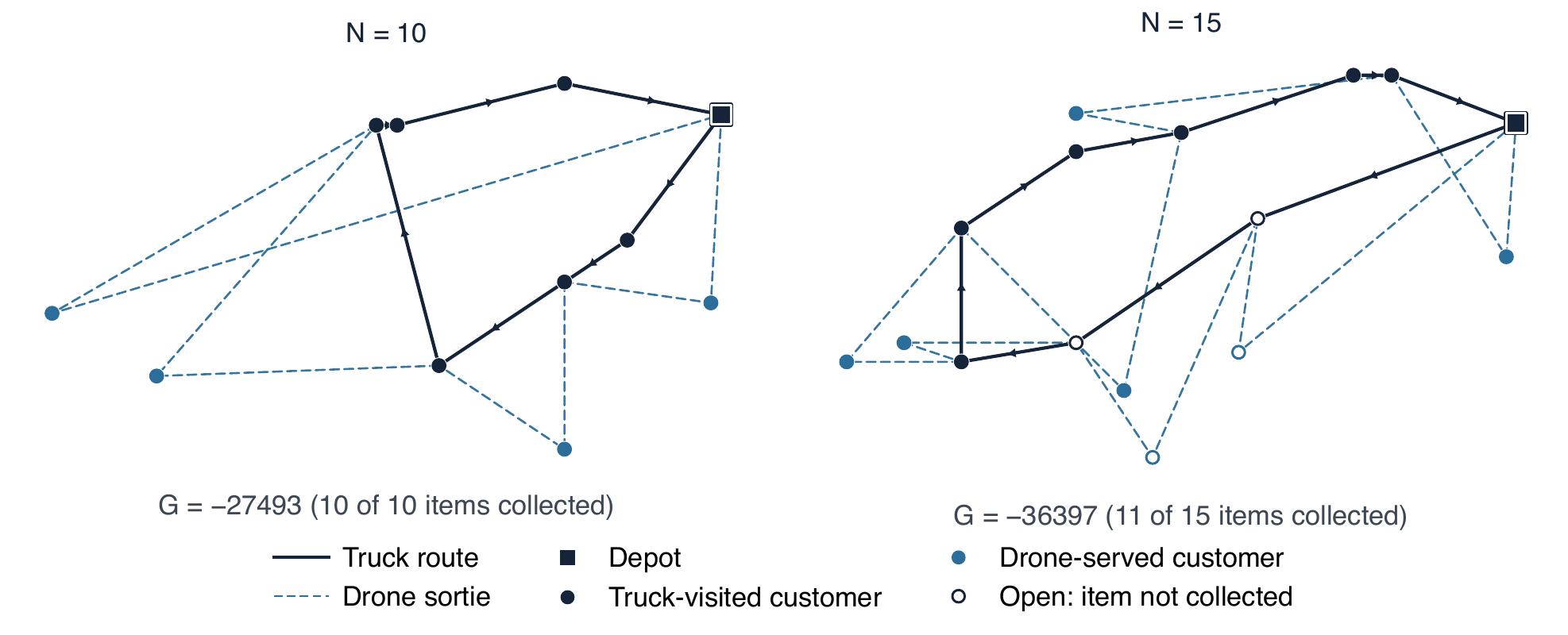}
\caption{TTP-D optimal plan on two instances with $N = 10$ and $N = 15$ customers. Each colour represents the visit mode. A filled circle marks a customer whose item is collected, and an open circle marks one who is visited but whose item is left behind.}
\label{fig:milp_routes}
\end{figure}

Figure~\ref{fig:ttpd_coupling} traces this coupling on a ten-customer instance. The horizontal axis is the mission time, the time elapsed since the fleet leaves the depot, whose value on the return to the depot is the makespan $\tau_{\depotend}$. Every item the truck collects adds weight and lowers its speed for the remainder of the route, so the closing arcs of the plan are driven at less than half the speed of the opening ones. The grey segments represent waiting times. In the first sortie, the drone reaches the rendezvous node before the truck and idles there, and in the second, the truck arrives first and is held at zero speed until the drone lands. Since the rental clock runs continuously, the launch and rendezvous nodes depend directly on the packing schedule.
Figure~\ref{fig:milp_routes} shows the optimal plans that this coupling produces on two small instances; Section~\ref{sec:exact_solver_results} returns to their structure.

The problem is stated under the following modelling assumptions.

\begin{itemize}
\item The problem is static and deterministic: all distances, profits, weights, and parameters are known before the mission begins and remain constant throughout its execution. A single truck and a single drone both start at the depot and must return to it. Each customer carries one collectable item, which may be collected only by the vehicle that visits that customer (Section~\ref{sec:ttd-multi-item} relaxes this to several items per customer).

\item The drone performs single-package sorties at a constant velocity $v_{\mathrm{D}}$, independent of its payload, and the item it carries cannot exceed $W^{\mathrm{D}}$. A sortie launches from a node at which the truck is present, flies to a single target customer, and terminates at a rendezvous node later on the truck's route, so it comprises exactly two drone arcs, one out and one back; the truck may visit customers of its own in between, and whichever vehicle reaches the rendezvous node first waits there for the other while the rental clock runs. Upon rendezvous, the drone-collected weight is transferred to the truck for onward transport. 

\item The truck operates with a finite knapsack capacity $W$, and it may not be feasible to collect all items. The truck's velocity is an affine function of its accumulated load, falling from $v_{\max}$ when empty to $v_{\min}$ at capacity, and its travel times are computed based on the departing weight at the start of an arc, rather than the arrival weight.

\item The transportation network relies on shared Euclidean inter-node distances for both vehicles, which are symmetric on both benchmarks; only the two speeds differ. Furthermore, operational delays, including loading, unloading, and inter-vehicle handovers, are considered negligible or are pre-absorbed into the travel distances.
\end{itemize}

\section{Mathematical Formulation}
\label{sec:MILP}

In this section, we present the mathematical formulation for the TTP-D. The proposed model is linear, except for the truck's travel time, which is inversely proportional to its load-dependent velocity $v$. This nonlinearity is convex with respect to the accumulated load $W_i$ and is addressed using Special Ordered Sets of type 2 (SOS2) Piecewise Linearisation (PL). Table~\ref{tab:notation} summarises the parameters and decision variables, grouped by their respective roles.

\begin{table}[h]
\centering
\caption{Notation for the TTP-D formulation.}
\label{tab:notation}
\small
\begin{tabularx}{\linewidth}{@{}llX@{}}
\toprule
\textbf{Symbol} & \textbf{Type} & \textbf{Meaning} \\
\midrule
\multicolumn{3}{@{}l}{\emph{Sets and indices}} \\
$\mathcal{N}$ & set & Customers, indexed $\mathcal{N} = \{1, \ldots, n\}$ \\
$N$ & $\N$ & Number of customers, $N = |\mathcal{N}|$  \\ 
$\mathcal{V}$ & set & Nodes, $\mathcal{V} = \{\depot, \depotend\} \cup \mathcal{N}$, the source and sink depots and the customers \\
$\mathcal{A}$ & set & Arcs, $\mathcal{A} = \{(i,j) : i,j \in \mathcal{V},\ i \neq j\}$ \\
$\delta^+(i), \delta^-(i)$ & set & Out-arcs and in-arcs at node $i$, respectively \\
$\delta^{\mathrm{D}}(j)$ & set & Drone in-arcs at $j$, $\{(k,j) \in \mathcal{A}: k \in \mathcal{N}\}$\\
$i, j$ & index & Generic nodes, $i, j \in \mathcal{V}$ \\
$k$ & index & Customer visited by a drone sortie, $k \in \mathcal{N}$ \\
$\ell$ & index & Node from which a drone sortie is launched \\
$b$ & index & Break-point of the PL, $b \in \{0, \ldots, K\}$ , where $K$ is the number of break-point intervals \\
\midrule
\multicolumn{3}{@{}l}{\emph{Instance Parameters}} \\
$d_{ij}$ & $\R_{\geq 0}$ & Euclidean distance from node $i$ to node $j$ \\
$d_{\max}$ & $\R_{>0}$ & $\max_{(i,j) \in \mathcal{A}} d_{ij}$ \\
$p_i$ & $\R_{\geq 0}$ & Profit of item at $i$; $p_{\depot} = p_{\depotend} = 0$ \\
$w_i$ & $\R_{\geq 0}$ & Weight of item at $i$; $w_{\depot} = w_{\depotend} = 0$ \\
$w_{\mathrm{tot}}$ & $\R_{\geq 0}$ & Total selectable weight, $w_{\mathrm{tot}} = \sum_{i \in \mathcal{N}} w_i$ \\
$W$ & $\R_{>0}$ & Truck capacity \\
$W^{\mathrm{D}}$ & $\R_{>0}$ & Drone payload capacity \\
$v_{\max}, v_{\min}$ & $\R_{>0}$ & Truck velocity at empty / full load, respectively; $v_{\min} \leq v_{\max}$ \\
$\Delta v$ & $\R_{\geq 0}$ & $\Delta v = v_{\max} - v_{\min}$ \\
$v_{\mathrm{D}}$ & $\R_{>0}$ & Constant drone velocity  \\
$R$ & $\R_{\geq 0}$ & Rental ratio per unit time \\
$K$ & $\N$ & Number of break-point intervals for PL \\
$w^{\mathrm{bp}}_b$ & $\R_{\geq 0}$ & Break-point weight, $w^{\mathrm{bp}}_b = b\,w_{\mathrm{tot}}/K$ for $b \in \{0, \ldots, K\}$ \\
$v^{\mathrm{bp}}_b$ & $\R_{>0}$ & Velocity at $w^{\mathrm{bp}}_b$, that is $v^{\mathrm{bp}}_b = v_{\max} - w^{\mathrm{bp}}_b \Delta v / W$ \\
\midrule
\multicolumn{3}{@{}l}{\emph{Decision variables}} \\
$x^{\mathrm{T}}_{ij}$ & $\{0,1\}$ & 1 if the truck traverses arc $(i,j)$ \\
$x^{\mathrm{D}}_{ij}$ & $\{0,1\}$ & 1 if the drone traverses arc $(i,j)$ \\
$y^{\mathrm{T}}_i$ & $\{0,1\}$ & 1 if customer $i$ is visited by the truck only \\
$y^{\mathrm{D}}_i$ & $\{0,1\}$ & 1 if customer $i$ is visited by the drone only \\
$y^{\mathrm{C}}_i$ & $\{0,1\}$ & 1 if customer $i$ is a rendezvous node; $y^{\mathrm{C}}_{\depot} = y^{\mathrm{C}}_{\depotend} = 1$ by convention \\
$z_i$ & $\{0,1\}$ & 1 if the item at $i$ is collected \\
$\lambda_{kj}$ & $[0,1]$ & Variable for McCormick linearisation of $z_k \cdot x^{\mathrm{D}}_{kj}$ \\
$W_i$ & $[0, W]$ & Truck weight on \emph{departure} from node $i$ \\
$\tau_i$ & $\R_{\geq 0}$ & Arrival time at $i$ (mission time); $\tau_{\depotend}$ is the makespan \\
$\hat\psi_i$ & $\R_{>0}$ & PL value at $i$ of the reciprocal truck speed $\psi$ \\
$\mu_{i,b}$ & $[0,1]\ \text{(SOS2)}$ & PL interpolation weight at $i$ for break-point $b$ \\
\bottomrule
\end{tabularx}
\end{table}

\subsection{Objective, Routing, and Assignment}

The objective function \eqref{eq:obj} maximises the total profit collected minus the rental cost, which is proportional to the makespan $\tau_{\depotend}$:
\begin{equation}
\max \quad G \;=\; \sum_{i \in \mathcal{N}} p_i \, z_i \;-\; R \cdot \tau_{\depotend}.
\label{eq:obj}
\end{equation}

Each customer $i \in \mathcal{N}$ is visited in exactly one mode (truck-only, drone-only, or truck at a rendezvous node):
\begin{equation}
y^{\mathrm{T}}_i + y^{\mathrm{D}}_i + y^{\mathrm{C}}_i \;=\; 1 \qquad \forall i \in \mathcal{N}.
\label{eq:partition}
\end{equation}

The truck routing is defined by flow conservation constraints \eqref{eq:truck-source-sink}--\eqref{eq:truck-flow}. In addition, constraints \eqref{eq:truck-degree} restrict truck visits to nodes designated either as truck-only or as rendezvous points:
\begin{align}
\sum_{j \,:\, (\depot,j) \in \mathcal{A}} x^{\mathrm{T}}_{\depot j} &= 1, \qquad \sum_{i \,:\, (i,\depotend) \in \mathcal{A}} x^{\mathrm{T}}_{i \depotend} = 1, \label{eq:truck-source-sink} \\
\sum_{j \,:\, (i,j) \in \mathcal{A}} x^{\mathrm{T}}_{ij} &= \sum_{j \,:\, (j,i) \in \mathcal{A}} x^{\mathrm{T}}_{ji} \qquad \forall i \in \mathcal{N}, \label{eq:truck-flow} \\
\sum_{j \,:\, (i,j) \in \mathcal{A}} x^{\mathrm{T}}_{ij} &= y^{\mathrm{T}}_i + y^{\mathrm{C}}_i \qquad \forall i \in \mathcal{N}. \label{eq:truck-degree}
\end{align}

Similarly, drone arcs consist of a launch leg from a rendezvous to a target and a return leg. Constraints \eqref{eq:drone-source-sink}–\eqref{eq:drone-out} enforce degree requirements based on the drone’s visit modes. In contrast, constraints \eqref{eq:drone-anchor} serve as an anchor condition, requiring each drone arc to visit at least one rendezvous node and thereby preventing inter-customer hopping.
\begin{align}
\sum_{j \,:\, (\depot,j) \in \mathcal{A}} x^{\mathrm{D}}_{\depot j} &= 1, \qquad \sum_{i \,:\, (i,\depotend) \in \mathcal{A}} x^{\mathrm{D}}_{i \depotend} = 1, \label{eq:drone-source-sink} \\
\sum_{j \,:\, (j,i) \in \mathcal{A}} x^{\mathrm{D}}_{ji} &= y^{\mathrm{D}}_i + y^{\mathrm{C}}_i \qquad \forall i \in \mathcal{N}, \label{eq:drone-in} \\
\sum_{j \,:\, (i,j) \in \mathcal{A}} x^{\mathrm{D}}_{ij} &= y^{\mathrm{D}}_i + y^{\mathrm{C}}_i \qquad \forall i \in \mathcal{N}, \label{eq:drone-out} 
\\
x^{\mathrm{D}}_{ij} &\;\leq\; y^{\mathrm{C}}_i + y^{\mathrm{C}}_j \qquad \forall (i,j) \in \mathcal{A}. \label{eq:drone-anchor}
\end{align}

Both depots act as anchors, because the truck is present at each. We therefore set $y^{\mathrm{C}}_{\depot} = y^{\mathrm{C}}_{\depotend} = 1$ by convention, so that \eqref{eq:drone-anchor} admits a sortie launched at the source depot and a sortie that rejoins the truck at the sink. These two values enter no other constraint, since the partition \eqref{eq:partition}, the truck degree \eqref{eq:truck-degree}, and the drone degrees \eqref{eq:drone-in}--\eqref{eq:drone-out} are stated over $\mathcal{N}$.

\subsection{Payload and Weight Tracking}

The drone may collect at most one item per sortie, strictly bounded by its payload limit $W^{\mathrm{D}}$:
\begin{equation}
w_i z_i \;\leq\; W^{\mathrm{D}} + M_{\mathrm{W}} (1 - y^{\mathrm{D}}_i) \qquad \forall i \in \mathcal{N}.
\label{eq:drone-payload}
\end{equation}

When $y^{\mathrm{D}}_i = 1$, constraints \eqref{eq:drone-payload} reduce to $w_i z_i \leq W^{\mathrm{D}}$, and otherwise it becomes redundant. Here, the big-$M$ constant is defined as $M_{\mathrm{W}} = \sum_i w_i$.
Along the active truck path, the departing weight at node $j$ equals the departing weight at the preceding node $i$, plus the weight of any item picked up at $j$, plus the weight of any items transferred from the drone upon rendezvous. To track the weight of drone-collected items (which requires the bilinear term $z_k \cdot x^{\mathrm{D}}_{kj}$ for a customer $k \in \mathcal{N}$ collected by the drone and delivered at the rendezvous node $j$), we introduce an auxiliary continuous variable $\lambda_{kj} \in [0,1]$. This product is linearised at all integer feasible points using standard McCormick envelopes:
\begin{equation}
\begin{aligned}
\lambda_{kj} &\;\geq\; z_k + x^{\mathrm{D}}_{kj} - 1, \\
\lambda_{kj} &\;\leq\; z_k, \\
\lambda_{kj} &\;\leq\; x^{\mathrm{D}}_{kj}, 
\end{aligned}
\qquad \forall\, (k,j) \in \mathcal{A}; \ k \in \mathcal{N}.
\label{eq:mccormick}
\end{equation}

\noindent Weight propagation is then enforced along each active truck arc $(i,j) \in \mathcal{A}$ using the linearised term:
\begin{align}
W_j &\;\geq\; W_i + w_j z_j + \sum_{(k,j) \in \delta^{\mathrm{D}}(j)} \lambda_{kj} w_k - M_{\mathrm{W}} (1 - x^{\mathrm{T}}_{ij}), \label{eq:wprop-lo}\\
W_j &\;\leq\; W_i + w_j z_j + \sum_{(k,j) \in \delta^{\mathrm{D}}(j)} \lambda_{kj} w_k + M_{\mathrm{W}} (1 - x^{\mathrm{T}}_{ij}), \label{eq:wprop-hi}
\end{align}
subject to the initial conditions $W_{\depot} = 0$, $z_{\depot} = z_{\depotend} = 0$, and bounds $0 \leq W_i \leq W$ for all $i \in \mathcal{N}$.

\subsection{Travel Time and Synchronisation}
\label{sec:sos2}

The truck's velocity is an affine function of its weight, $v_i = v_{\max} - W_i \Delta v / W$. The time spent traversing arc $(i,j)$ relies on the reciprocal travel time function $\psi(W_i):= 1/v_i$. Since $v_i$ is affine on $[0, W]$, $\psi(W_i)$ is strictly convex and increasing. We approximate this non-linearity using a chord interpolant over $K$ uniform breakpoints distributed across the realisable load range $[0, w_{\mathrm{tot}}]$. For each node $i \in \mathcal{V}$, we introduce nonnegative weights $\mu_{i,b}$ bounded by SOS2 condition:
\begin{align}
W_i &\;=\; \sum_{b=0}^{K} \mu_{i,b} \, w^{\mathrm{bp}}_b, \qquad \hat\psi_i \;=\; \sum_{b=0}^{K} \mu_{i,b} \, \frac{1}{v^{\mathrm{bp}}_b}, \label{eq:sos2-w}\\
\sum_{b=0}^{K} \mu_{i,b} &\;=\; 1, \qquad \mu_{i,b} \geq 0, \qquad \{\mu_{i,b}\}_{b=0}^{K} \text{ is SOS2.} \label{eq:sos2-conv}
\end{align}
The travel time over an active truck arc is bounded by $d_{ij} \hat\psi_i$, while drone timing assumes a constant velocity $v_{\mathrm{D}}$ on active flight segments:
\begin{align}
\tau_i + d_{ij} \, \hat\psi_i \;&\leq\; \tau_j + M_{\mathrm{T}} (1 - x^{\mathrm{T}}_{ij}) \qquad \forall (i,j) \in \mathcal{A}, \label{eq:truck-timing} \\
\tau_i + \frac{d_{ij}}{v_{\mathrm{D}}} \;&\leq\; \tau_j + M_{\mathrm{D}} \left( 1 - x^{\mathrm{D}}_{ij} + x^{\mathrm{T}}_{ij} \right) \qquad \forall (i,j) \in \mathcal{A}. \label{eq:drone-timing}
\end{align}
In Eq.~\eqref{eq:drone-timing}, the multiplier $(1 - x^{\mathrm{D}}_{ij} + x^{\mathrm{T}}_{ij})$ ensures flight time is only calculated when the drone flies independently of the truck. 
Each disjunctive constant is set to its smallest valid value: $M_{\mathrm{W}} = w_{\mathrm{tot}}$ in \eqref{eq:drone-payload} and \eqref{eq:wprop-lo}--\eqref{eq:wprop-hi}, $M_{\mathrm{T}} = d_{\max}(N+1)/v_{\min}$ for \eqref{eq:truck-timing}, and $M_{\mathrm{D}} = d_{\mathrm{NN}}/v_{\min}$ for \eqref{eq:drone-timing}, where $d_{\mathrm{NN}}$ denotes the length of the nearest-neighbour warm-start tour. Table~S1 in the Supplementary Material provides a detailed explanation of each bound.

$\psi$ is convex, and the SOS2 chord strictly bounds the true curve from above ($\hat\psi_i \geq 1/v_i$). This guarantees a conservative approximation ($G^*_{\mathrm{MILP}} \leq G^*_{\mathrm{exact}}$) where the maximum objective error propagated through the path length is bounded by:
\begin{equation}
\epsilon_G \;\leq\; R \cdot d_{\max} \cdot (N+1) \cdot \left(\frac{w_{\mathrm{tot}}}{W}\right)^{\!2} \frac{(\Delta v)^2}{4 K^2 v_{\min}^3}.
\label{eq:obj-error}
\end{equation}
Section~S2 of the Supplementary Material derives \eqref{eq:obj-error} and states the two-sided form $0 \leq G^*_{\mathrm{exact}} - G^*_{\mathrm{MILP}} \leq \epsilon_G$.

\subsection{Valid Inequalities}
\label{sec:valid-ineq}

The per-arc timing constraints leave the linear relaxation of $\tau_{\depotend}$ weak. We introduce three valid inequalities to directly bound the makespan from below. Constraints \eqref{eq:vi-truck} and \eqref{eq:vi-drone} aggregate the minimum required travel time based on the total distance covered by each vehicle:
\begin{align}
\tau_{\depotend} &\;\geq\; \frac{1}{v_{\max}} \sum_{(i,j) \in \mathcal{A}} d_{ij}\, x^{\mathrm{T}}_{ij}, \label{eq:vi-truck}\\
\tau_{\depotend} &\;\geq\; \frac{1}{v_{\mathrm{D}}} \sum_{(i,j) \in \mathcal{A}} d_{ij}\, x^{\mathrm{D}}_{ij}, \label{eq:vi-drone} \\
\tau_{\depotend} &\;\geq\; \big(d_{\depot i} + d_{i \depotend}\big)
   \left( \frac{y^{\mathrm{T}}_i + y^{\mathrm{C}}_i}{v_{\max}} + \frac{y^{\mathrm{D}}_i}{v_{\mathrm{D}}} \right)
   \qquad \forall i \in \mathcal{N}. \label{eq:vi-roundtrip}
\end{align}

Constraints \eqref{eq:vi-roundtrip} enforce a mode-aware round-trip minimum for each customer $i$. Every customer must be visited, meaning the vehicle visiting $i$ traverses at least the direct distance from the depot to $i$ and back. The inequality establishes this minimum by bounding the spatial distance by the maximum speed of the assigned mode: $v_{\max}$ for the truck and $v_{\mathrm{D}}$ for the drone. 
For drone-visited customers, this bound holds via the triangle inequality across the truck-drone path, provided $v_{\mathrm{D}} \geq v_{\max}$. These valid inequalities tighten the root LP bound by up to fivefold in our computational experiments on instances with $N=20$, serving as the driver in closing the $N=15$ instances to optimality.

\section{Solution Methods}
\label{sec:methodology}

While the MILP formulation guarantees optimality, it is tractable only for small-scale instances. This section introduces three scalable approaches in order of increasing sophistication. 
Section~\ref{sec:metaheuristics} presents metaheuristic baselines that search the joint route–pack–sortie space anew for each instance. 
Section~\ref{sec:drl} shifts the computational cost of per-instance search to an offline training phase, using a reinforcement learning policy that constructs plans in a single forward pass. 
Finally, Section~\ref{sec:LISA} introduces LISA. This hybrid approach integrates the metaheuristic into the policy and uses a minimal search budget to repair generated plans, trading a marginal reduction in solution quality for substantial time savings.
Table~\ref{tab:notation-methods} summarises the notation used in the solution methods.

\begin{table}[tbh]
\centering
\caption{Notation for the solution methods.}
\label{tab:notation-methods}
\small
\begin{tabularx}{\linewidth}{@{}llX@{}}
\toprule
\textbf{Symbol} & \textbf{Type} & \textbf{Meaning} \\
\midrule
\multicolumn{3}{@{}l}{\emph{Metaheuristics (Section~\ref{sec:metaheuristics})}} \\
$\mathbf{x}, \mathbf{x}'$ & tuple & Incumbent solution $\mathbf{x} = (\mathbf{r}, \mathcal{S}, \mathbf{z})$, and a feasible neighbour $\mathbf{x}'$ \\
$\mathbf{r}$ & vector & Order in which the truck visits its customers \\
$\mathcal{S}$ & set & Drone sorties, each defined as a triple $(\ell, k, j)$ \\
$\mathbf{z}$ & vector & Packing configuration, with binary components $z_i$ \\
$\Delta G$ & $\R$ & Change in objective function value from a move; $G(\mathbf{x}') - G(\mathbf{x})$ \\
$T, T_0$ & $\R_{>0}$ & Annealing temperature and its calibrated initial value, respectively \\
$\eta$ & $\N$ & VNS perturbation strength, where $1 \leq \eta \leq \eta_{\max}$ \\
$B(N)$ & $\R_{>0}$ & Instance-size-dependent runtime budget \\
\midrule
\multicolumn{3}{@{}l}{\emph{Markov decision process (Section~\ref{sec:drl-mdp})}} \\
$t$ & index & Decision epoch \\
$s_t, a_t, r_t$ & --- & State, composite action, and step reward at epoch $t$ \\
$c_t$ & $\mathcal{V}$ & Truck's current node \\
$\tau_t$ & $\R_{\geq 0}$ & Elapsed mission time upon arrival at node $c_t$ \\
$W_t, W'_t$ & $[0, W]$ & Truck load upon arrival at and departure from node $c_t$, respectively \\
$\mathcal{V}_t$ & set & Subset of customers visited by epoch $t$; $\mathcal{V}_t \subseteq \mathcal{N}$ \\
$F_t$ & --- & Drone status: idle, or in-flight from $\ell$ toward $k$ \\
$\tau^{\mathrm{D}}$ & $\R_{\geq 0}$ & Drone's arrival time at the rendezvous node \\
$\rho_t$ & $\{0,1\}$ & Binary decision to land the in-flight drone at $c_t$ \\
$z^{\mathrm{D}}_t, z^{\mathrm{T}}_t$ & $\{0,1\}$ & Binary decisions to collect the drone's item and the item at $c_t$, respectively \\
$k_t$ & $\mathcal{N} \cup \{\varnothing\}$ & Drone launch target, with $\varnothing$ representing an idle state \\
$j_t$ & $\mathcal{N} \cup \{\depotend\}$ & Target node the truck drives to next \\
\midrule
\multicolumn{3}{@{}l}{\emph{Policy and training (Sections~\ref{sec:drl-policy}--\ref{sec:drl-training})}} \\
$I$ & instance & Training instance drawn from the benchmark distribution \\
$\pi_\theta$ & policy & Learned construction policy parametrised by weights $\theta$ \\
$h_i$ & $\R^{d_{\mathrm{model}}}$ & Learned feature embedding of node $i$ \\
$d_{\mathrm{model}}$ & $\N$ & Dimensionality of the neural network model ($128$) \\
$P$ & $\N$ & Number of rollouts per instance (POMO group size) \\
$m$ & index & Rollout within a group, $m = 1, \ldots, P$ \\
$G^{(m)}$ & $\R$ & Total episode return achieved by rollout $m$ \\
$A_t$ & $\R$ & Advantage estimate at step $t$, group-centred and $\kappa_I$-scaled \\
$\kappa_I$ & $\R_{>0}$ & Stationary per-instance reward scaling factor \\
$\omega_t(\theta)$ & $\R_{>0}$ & PPO probability ratio at step $t$ \\
$\epsilon_{\mathrm{clip}}$ & $\R_{>0}$ & PPO clipping parameter \\
$\zeta$ & $\R_{\geq 0}$ & Entropy regularisation coefficient \\
$\mathcal{L}(\theta)$ & $\R$ & Clipped surrogate objective \\
\midrule
\multicolumn{3}{@{}l}{\emph{LISA (Section~\ref{sec:LISA})}} \\
$\beta$ & $(0,1]$ & Fraction of $B(N)$ allocated to the annealing repair \\
\bottomrule
\end{tabularx}
\end{table}

\subsection{Metaheuristic Solvers}
\label{sec:metaheuristics}

Heuristics navigate the highly coupled TTP-D by exploring the joint route--pack--sortie space directly, foregoing the exhaustive search required by exact solvers to guarantee optimality. Simple constructive approaches, however, are inherently limited. 
Consider a greedy strategy that constructs a nearest-neighbour tour from the depot, packs items by profit-to-weight ratio, and assigns drone sorties via best insertion. Once the tour and packing plan are fixed, no best-insertion sortie can rectify a suboptimal visiting sequence, whilst an early, heavy pickup incurs a cumulative travel-time penalty across every subsequent arc. In our preliminary experiments, such greedy constructions over-committed weight and incurred large rental penalties.  Metaheuristics overcome this rigidity by iteratively revising prior decisions. To this end, we propose two metaheuristic algorithms. Both warm-start from the nearest-neighbour solution, draw candidate moves from a shared move library, employ an identical feasibility-preserving evaluator, and operate within a common, instance-size-dependent runtime budget.

\paragraph{Move library} A solution comprises a truck visitation sequence, a set of drone sorties, each specified by a launch node, a target customer, and a rejoin node, and a binary collection decision per customer. Ten operators systematically alter these components. They comprise a collection flip; three route operators (truck-customer swaps, $2$-opt reversals, and Or-opt relocations of segments of length one to three); two sortie operators that preserve the truck route (exact re-scheduling of the launch and rejoin anchors by dynamic programming over subsets of the drone targets, applied up to a fixed sortie count, and re-anchoring of a single sortie over all admissible launch–rejoin pairs); and four transfer operators that move customers between the truck and the drone by exchanging a truck customer with a drone target, returning a drone target to the truck, or promoting a truck customer to a drone target either under every admissible anchor pair or with all sorties re-scheduled exactly. If a route operator inverts a sortie's launch and rejoin nodes, an anchor repair swaps them back. During evaluation, infeasible moves such as overlapping sorties, disconnected anchors, or capacity violations are strictly pruned.

\paragraph{Search strategies} The move library supports a Variable Neighbourhood Descent (VND) that cycles through the operators under a first-improvement rule, restarting upon finding an improving move. For larger $N$, each pass evaluates at most $30N$ randomly sampled candidates per operator, since exhaustively searching the $O(N^2)$-$O(N^3)$ neighbourhood spaces is computationally expensive.
The metaheuristics utilise this framework: SA samples candidate moves globally, reserving the VND solely for the warm start and final polishing, whereas VNS uses the sampler for stochastic shaking and the VND for local descent.

\subsubsection{Simulated Annealing}
\label{sec:sa}

Simulated annealing \citep{kirkpatrick1983optimization} is a single-trajectory metaheuristic that accepts worsening moves with a temperature-controlled probability, allowing the search to escape local optima and continue exploring the solution space. We follow the cooling-schedule design that has proven effective for the classical TTP \citep{elyafrani2018efficiently}. Since $G$ is a return to be maximised, a move from $\mathbf{x}$ to a feasible neighbour $\mathbf{x}'$ with $\Delta G = G(\mathbf{x}') - G(\mathbf{x})$ is accepted with probability
\begin{equation}
\Pr(\text{accept} \given \Delta G, T) =
\begin{cases}
1 & \Delta G \geq 0,\\[2pt]
\exp(\Delta G / T) & \Delta G < 0.
\end{cases}
\label{eq:metropolis}
\end{equation}
So, improving moves are always taken and worsening moves pass with a probability that falls as the temperature $T$ cools.

We initialise the search with the nearest-neighbour solution and refine it using a single VND descent. The initial temperature is calibrated to the instance by sampling worsening moves from this starting point and setting $T_0 = -\overline{|\Delta G^-|} / \ln 0.8$, where $\overline{|\Delta G^-|}$ is the mean magnitude of the sampled worsening changes, so that roughly $80\%$ of worsening moves are accepted at the outset. Each temperature level proposes $\max(20,\, 12(N+1))$ random feasible neighbours from the sampler of Section~\ref{sec:metaheuristics}, after which the temperature cools geometrically, $T \leftarrow \alpha T$ with $\alpha = 0.97$. When $T$ falls below $10^{-4} T_0$, the schedule reheats to $T_0$ and resumes from the incumbent best, and a final descent polishes the best solution found.

\subsubsection{Variable Neighbourhood Search}
\label{sec:vns}

Variable neighbourhood search \citep{mladenovic1997variable, hansen2001variable} systematically varies the neighbourhood structure during the search, alternating between a stochastic shaking phase to escape local optima and a deterministic local descent to re-optimise the incumbent solution. Since preliminary testing of a General VNS variant that restarts the search after stagnation yielded identical results, we adopt the standard VNS for simplicity. 
The VNS shakes the incumbent solution using $\eta$ random moves generated by the sampler described in Section~\ref{sec:metaheuristics}, and then applies the VND. The resulting solution is accepted only if it improves upon the incumbent. Upon success, the search recentres on the new incumbent and resets the perturbation strength to $\eta = 1$. Otherwise, it increments the perturbation strength to $\eta + 1$, cycling back to $1$ once it reaches $\eta_{\max} = 8$. 

\subsection{Deep Reinforcement Learning Solver}
\label{sec:drl}

While metaheuristics solve every instance independently without transferring learned information, we learn a construction policy that amortises search effort across instances. Following the encoder–decoder paradigm established for routing problems by \citet{vinyals2015pointer} and refined by \citet{bello2016neural} and \citet{kool2019attention}, the policy constructs feasible TTP-D solutions sequentially. The encoder embeds the instance once, after which the decoder incrementally extends partial solutions by conditioning each decision on both the instance embedding and the current solution state. Solving a new instance thus requires only a forward pass per decision, with no runtime-dependent search trajectory. Since training is performed offline only once, the learned policy generalises to new instances drawn from the same distribution without retraining, enabling fast inference suitable for operational settings that require frequent route re-planning. Coupled routing, packing, and sortie decisions are emitted in a fixed sequence via a factored action space, with feasibility enforced by action masking to guarantee solutions admissible under Section~\ref{sec:MILP}.

Two design choices govern policy training. First, we optimise the policy using Proximal Policy Optimisation (PPO; \citealp{schulman2017proximal}), whose clipped surrogate objective safely reuses rollout batches across multiple gradient epochs while constraining updates. Second, we adopt the multi-start group baseline introduced by POMO \citep{kwon2020pomo}. Recognising that routing instances admit equivalent solutions that differ only in their initial decisions, POMO rolls out the policy from multiple forced first moves and uses the group-mean return as an instance-specific baseline. This eliminates the need for a learned critic, avoiding additional parameters and potential estimation bias, while providing structured exploration. In the TTP-D, the initial truck movement and drone launch choices naturally provide these distinct starting points across the factored action space.

\subsubsection{Markov Decision Process}
\label{sec:drl-mdp}

We formulate the TTP-D as a finite-horizon, deterministic Markov Decision Process (MDP) in which every episode constructs a feasible solution to the MILP. The environment operates as a discrete-event simulator that tracks the truck. Decision epochs occur whenever the truck reaches a node, aggregating all available choices into a single composite action. We utilise the instance parameters defined in Table~\ref{tab:notation}.

\paragraph{State} After $t$ epochs, the state is represented as:
\begin{equation}
s_t = \left(c_t,\ \tau_t,\ W_t,\ \mathcal{V}_t,\ F_t\right),
\end{equation}
where $c_t \in \mathcal{V}$ is the truck's current node (with $0$ the source depot \depot\ and $N+1$ the sink), $\tau_t \in \R_{\geq 0}$ is the elapsed mission time, $W_t \in [0, W]$ is the truck load upon arrival at $c_t$, and $\mathcal{V}_t \subseteq \mathcal{N}$ is the set of customers already visited. The drone's status $F_t$ indicates whether it is idle or in flight. If in flight, $F_t$ records the launch node $\ell$, the target customer $k$, and the launch time. Because this tuple fully determines the drone's arrival at any future rendezvous, the Markov property is preserved. The policy also reads two timing quantities from $s_t$: the wait time, $\max\{0,\ \tau^{\mathrm{D}} - \tau_t\}$, if the drone were to land at $c_t$, and the airborne time, $\tau_t - \tau_{\mathrm{launch}}$.

\paragraph{Composite action} A composite action is defined by the tuple
\begin{equation}
a_t = \left(\rho_t,\ z^{\mathrm{D}}_t,\ z^{\mathrm{T}}_t,\ k_t,\ j_t\right),
\end{equation}
which is decoded autoregressively in the following order:
\begin{enumerate}\itemsep2pt
  \item $\rho_t \in \{0, 1\}$: Dictates whether the in-flight drone rejoins the truck at $c_t$. This decision is forced to 1 at the sink if the drone is not on the truck, and fixed to 0 if the drone is idle.
  \item $z^{\mathrm{D}}_t \in \{0, 1\}$:  Determines whether to collect the item delivered by the landing drone. This is used only when $\rho_t = 1$, and is masked by the truck's remaining capacity, as the payload transfers to the truck upon landing.
  \item $z^{\mathrm{T}}_t \in \{0, 1\}$: Determines whether to collect the item at the current node $c_t$, masked by
    capacity.
  \item $k_t \in \mathcal{N} \cup \{\varnothing\}$: Selects the drone's launch target, available only if the drone remains idle. The target $k_t$ must be unvisited, uncommitted, and within the drone's payload limit. The null action $\varnothing$ keeps the drone idle.
  \item $j_t \in \mathcal{N} \cup \{\depotend\}$: Selects the next node for the truck to visit from the pool of unvisited and uncommitted customers. The sink \depotend\ becomes admissible only after all customers have been visited.
\end{enumerate}
The launch decision deliberately precedes the truck move, making the latter conditional on the launch outcome. For instance, reversing this sequence precludes scenarios in which the drone is launched to visit the final customer while the truck proceeds directly to the depot to await the drone's return.

\paragraph{Transition} The simulator executes the action based on the policy. Upon a rejoin, the mission time advances to $\max(\tau_t, \tau^{\mathrm{D}})$, where $\tau^{\mathrm{D}}$ is the drone arrival time, ensuring the earlier arrival waits for the later one. Any collected drone payload is then transferred to the truck's load. The truck subsequently picks up the item at $c_t$ if $z^{\mathrm{T}}_t = 1$, resulting in a departing load of 
$W'_t = W_t + w_{c_t}\, z^{\mathrm{T}}_t$.
Additionally, a launched drone visits exactly one customer before its next rendezvous, consistent with the MILP constraints. The truck then travels to $j_t$ at a load-dependent speed, advancing the clock by the corresponding travel time:
\begin{equation}
v(W'_t) = v_{\max} - W'_t\,\frac{v_{\max} - v_{\min}}{W},
\qquad
\tau_{t+1} = \tau_t + \frac{d_{c_t j_t}}{v(W'_t)},
\end{equation}
and $j_t$ is marked as visited. For a drone in flight from $\ell$ to $k$ and rejoining at $c_t$, its arrival time is calculated as $\tau^{\mathrm{D}} = \tau_{\mathrm{launch}} + (d_{\ell k} + d_{k c_t})/v_{\mathrm{D}}$.

\paragraph{Reward and objective} The step reward is the profit collected at the epoch minus the rental cost of the elapsed time,
\begin{equation}
r_t = \left(p_{c_t} z^{\mathrm{T}}_t + p_{k} z^{\mathrm{D}}_t\right) - R\,(\tau_{t+1} - \tau_t),
\end{equation}
where $k$ is the customer the landing drone visited. Without discounting ($\gamma = 1$), the rental costs telescope, making the total episode return equivalent to the TTP-D objective:
\begin{equation}
\sum_t r_t = \sum_{i \in \mathcal{N}} p_i z_i - R\,\tau_{\depotend} = G.
\end{equation}
An episode terminates when the truck reaches the sink, all customers are visited, and the drone has landed. To improve training, we truncate and fail episodes after $2(N+2)$ epochs, a threshold no valid sequence reaches.

\paragraph{Feasibility} Each action head is decoded under a feasibility mask that prevents invalid choices, including capacity violations for both collection bits, overlapping commitments for $k_t$ and $j_t$, and premature sink visits. These masks are bidirectional: every admissible composite action transitions to a state with at least one feasible completion, eliminating the need for backtracking or post-hoc repairs. 
Conversely, every feasible MILP solution maps to a valid action sequence. We validated the simulator by replaying the MILP solver's optimal solutions.

\subsubsection{Policy Network}
\label{sec:drl-policy}

Our policy network adopts the encoder-decoder architecture of \citet{kool2019attention}. 
A permutation-equivariant attention encoder embeds the problem instance once per episode, after which a lightweight decoder generates a distribution for each action head at every epoch, conditioned on the node embeddings and the current state context.

\paragraph{Encoder} Each of the $N+2$ nodes (the source, the $N$ customers, and the sink copy of the depot) is represented by four features: two coordinates normalised to the unit square using the instance bounding box, the profit normalised by the maximum profit in the instance, and the weight normalised by the truck capacity. A linear map projects the node features into the model dimension, followed by a stack of pre-norm multi-head self-attention blocks, each with a position-wise feed-forward network, residual connections, and layer normalisation. Detailed hyperparameters are provided in Table~S2 of the Supplementary Material. A final layer normalisation produces the node embeddings $h_1, \ldots, h_{N+2}$. Attention is computed over the complete node set, and the encoder instantiates the graph-attention inductive bias \citep{velickovic2018graph} on a fully connected instance graph. The encoder is evaluated once per episode and once per augmentation view (Section~\ref{sec:drl-training}), with its output reused across all subsequent decoding steps. 

\paragraph{State context} At each decision step, the decoder forms a context vector by concatenating the mean node embedding, the embedding of the current node $c_t$, the mean embedding of the unvisited customers, and seven state features: the load fraction $W_t/W$, the elapsed time normalised by the tour-time bound $d_{\max}(N+1)/v_{\max}$, the fraction of customers remaining, the fraction of total profit collected, an in-flight indicator, and the two normalised drone timing features described above. On \texttt{ttd300}, where the drone carries a finite endurance, two further features, normalised endurance and in-flight slack, are appended and the count rises to nine (Section~\ref{sec:ttd300-bench}). A two-layer Gaussian Error Linear Unit (GELU) multilayer perceptron maps the resulting vector to a $d_{\mathrm{model}}$-dimensional context shared across all action heads.

\paragraph{Action heads} The two node-selection heads, the launch target $k_t$ and the truck move $j_t$, are pointer heads \citep{vinyals2015pointer, kool2019attention}. Specifically, an $8$-head attention glimpse refines the context query over the masked candidate embeddings. 
This is followed by a single-head compatibility score scaled by $\sqrt{d_{\mathrm{model}}}$, which is then clipped to $[-C, C]$ via $C\tanh(\cdot)$ using $C = 10$; masked entries are assigned $-\infty$ before the softmax operation. The candidate set for the $k_t$ head is the node embeddings augmented by a single learned dummy embedding representing ``no launch''. The $j_t$ head factors in the launch decision by concatenating the chosen target's embedding, or the dummy, to the context query. 
The three binary decision heads ($\rho_t$, $z^{\mathrm{D}}_t$, and $z^{\mathrm{T}}_t$) are implemented as two-layer GELU multilayer perceptrons. Each processes the context vector concatenated with the most relevant node embedding: the in-flight target is used for $\rho_t$ and $z^{\mathrm{D}}_t$, while the current node is used for $z^{\mathrm{T}}_t$. 

The joint policy at any given decision step factorises into the product of the five individual head distributions. Each distribution is conditioned on the outcomes of the preceding heads in the decoding sequence:
\begin{equation}
\pi_\theta(a_t \given s_t)
  = \pi^{\rho}_\theta\,\pi^{z^{\mathrm{D}}}_\theta\,\pi^{z^{\mathrm{T}}}_\theta\,\pi^{k}_\theta\,\pi^{j}_\theta.
\end{equation}
Consequently, the log-probability of a composite action equates to the sum of the log-probabilities of the five constituent heads, and the total entropy regulariser is the sum of their respective entropies.

\paragraph{Encoder Ablation}
\label{sec:drl-ablation}
To isolate the contribution of inter-node information exchange, we replace each graph attention sub-layer with a node-wise residual feed-forward block of identical width, so that each node’s embedding is computed solely from its own four input features. Everything else is held fixed: the encoder depth, embedding dimension, feed-forward width, decoder, five action heads, action masks, PPO training procedure, POMO baseline, and training schedule. The resulting node-wise multilayer perceptron (MLP) policy is trained from scratch, and the comparative results are presented in Section~\ref{sec:a280-comparison}.

\subsubsection{Training}
\label{sec:drl-training}

The policy is trained via PPO to maximise the episode return $G$. Instead of a learned critic, we employ a POMO-based group baseline. As the action space is factored across five heads, we adapt three core components of the training procedure: the baseline formulation, the clipped PPO update, and POMO's forced-diverse starts.

\paragraph{Group baseline}
For each instance $I$, we execute $P = 32$ rollouts of the current policy and use the group's mean return as an instance-conditional baseline \citep{kwon2020pomo}. The advantage assigned to every step of rollout $m$ is the centred return:
\begin{equation}
A^{(m)} = \frac{1}{\kappa_I}
  \left(G^{(m)} - \frac{1}{P}\sum_{m'=1}^{P} G^{(m')}\right),
\qquad
\kappa_I = \max\!\left(1,\ \frac{R\,d_{\max}(N+1)}{v_{\max}}\right),
\end{equation}
where $\kappa_I$ is a per-instance stationary scale. 
This normalises reward magnitudes, which vary across instance sizes (e.g., from approximately $-17{,}000$ at $N = 5$ to $-40{,}000$ at $N = 20$). Advantages are standardised across the training batch. 

\paragraph{PPO update} Let
$\omega_t(\theta) = \pi_\theta(a_t \given s_t)/\pi_{\theta_{\mathrm{old}}}(a_t \given s_t)$
be the ratio of the stored composite action at step $t$ under the current policy relative to the behaviour policy. The policy maximises the clipped surrogate:
\begin{equation}
\mathcal{L}(\theta) = \mathbb{E}_t\!\left[
  \min\!\big(\omega_t(\theta)\,A_t,\ \operatorname{clip}(\omega_t(\theta), 1-\epsilon_{\mathrm{clip}}, 1+\epsilon_{\mathrm{clip}})\,A_t\big)
  \right] + \zeta\, H[\pi_\theta],
\end{equation}
where $\epsilon_{\mathrm{clip}}$ restricts the magnitude of policy updates and the entropy coefficient $\zeta$ is annealed on a cosine schedule, so that exploration is broad early and the policy sharpens as training proceeds. 
Rollout batches are reused across several update epochs in mini-batches. To prevent excessive deviation from the behaviour policy, an update epoch stops early if the mean approximate KL divergence exceeds a predefined target. Gradients propagate through the encoder, which is re-evaluated via automatic differentiation within each mini-batch, and the global gradient norm is clipped. We use the AdamW optimiser with a warmup-cosine learning-rate schedule.
The hyperparameter values are provided in Table S2 of the Supplementary Material.

\paragraph{Start strata, pinned heads, and augmentation} POMO necessitates forced-diverse initial moves. Given our factored action space, we stratify these starts jointly over the first truck move $j_0$ (spanning all feasible initial nodes) and the first drone launch $k_0$ (covering all depot launch options, including the dummy action). The $P$ rollouts are assigned $(j_0, k_0)$ pairs by cycling through both lists using decoupled strides, thereby avoiding a rigid product ordering. Furthermore, each rollout is subjected to one of eight dihedral transformations (four rotations and their reflections) applied to the normalised coordinates \citep{kwon2020pomo}. The encoder processes each of these augmentation views exactly once, sharing the resulting eight encodings across all $P$ rollouts. Pinned initial decisions are excluded from both the behaviour and gradient-time log-probabilities, ensuring $\omega_t$ is computed exclusively over policy-selected actions. If the first truck move is pinned while the launch head remains free, the pinned node is subsequently masked from the launch options. All rollouts within a group progress in lockstep via a single batched forward pass per decoding step. 

\subsubsection{Inference}
\label{sec:drl-inference}

During inference, the policy functions as a solution sampler, from which the best feasible rollout is selected. Across both benchmarks, solutions are decoded using a beam search of width $256$ under all eight dihedral views, serving as the inference-time equivalent of the training strata.
The beam expands the prefixes with the highest cumulative log-probability, keeping one beam per first-launch option, and the best completed solution is selected by the objective \citep{joshi2019efficient, choo2022simulation}. Each reported plan is replayed in the simulator and verified as feasible before scoring. This beam search configuration was selected following an ablation comparing seven candidate decoding strategies, ranging from simple greedy rollouts and multi-start variants to sampling and search-augmented decoding. A detailed description and performance comparison of these strategies is provided in Section~S3 of the Supplementary Material, while Table~S2 lists the remaining hyperparameters.

\subsection{Learner-Initialised Simulated Annealing}
\label{sec:LISA}

The DRL and metaheuristic solver families offer complementary trade-offs. While the learned policy amortises search costs via offline training to enable rapid, single-pass inference, its performance gap often widens as the problem scale increases. SA, by contrast, reaches the best objectives we observe at every size, but requires a full optimisation run for each instance. Learner-Initialised Simulated Annealing (LISA) bridges these approaches. By distilling the metaheuristic's search behaviour into a neural policy, LISA generates a robust initial solution, which is then refined through a truncated SA phase, thereby reducing online computation to a low-budget local search.

\begin{figure}[htbp]
\centering
\includegraphics[width=0.8\textwidth]{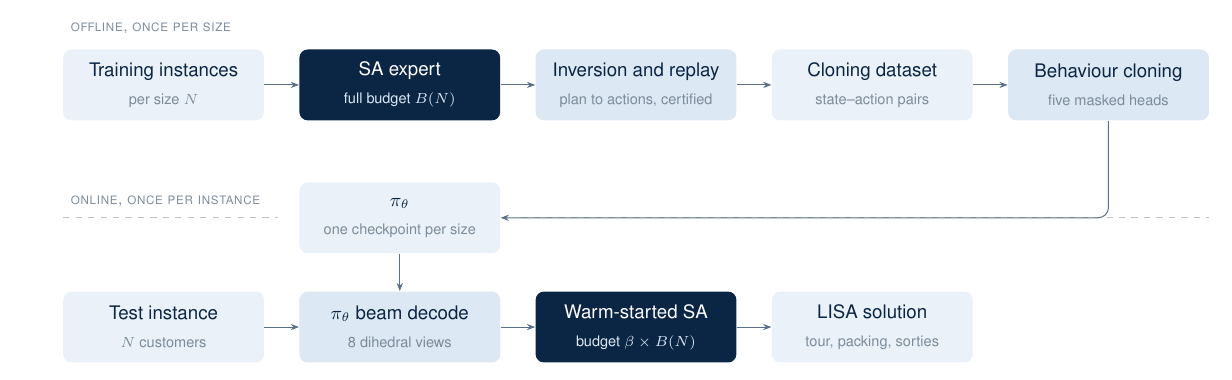}
\caption{The LISA framework. The upper track depicts offline policy distillation, while the lower track depicts online inference, where the cloned policy $\pi_\theta$ warm-starts SA.}
\label{fig:lisa_arch}
\end{figure}

LISA is built in three stages. First, the full-budget SA solver is applied to a corpus of training instances. Each expert solution is then converted by an inverter into the composite action sequence of the MDP. The resulting sequence is replayed in the simulator and verified to reproduce the expert objective before being added to the training dataset. Second, the attention policy is trained on the certified state–action pairs via behaviour cloning, using supervised maximum-likelihood estimation over the five factored action heads, under the same feasibility masks employed during DRL. A distinct policy checkpoint is trained for each instance size. Finally, during inference, the cloned policy decodes a beam of candidate plans (beam width 128, eight dihedral views). The best feasible plan then warm-starts an SA run governed by a reduced runtime budget of $\beta B(N)$, where $\beta \in (0,1]$. As illustrated in Figure~\ref{fig:lisa_arch}, the upper track denotes the offline distillation of the SA expert into the policy $\pi_\theta$, while the lower track represents the online inference phase. 

The single parameter $\beta$ makes LISA an anytime solver that unifies policy inference with SA. As $\beta \rightarrow 0$, it reduces to the cloned policy; at $\beta = 1$, it recovers the full-budget SA solver. Sections ~{\ref{sec:a280-comparison}} and~\ref{sec:ttd300-results} evaluate this trade-off on both benchmarks using budgets from $5\%$ to $50\%$ of the full annealing budget.

\section{Computational Study}
\label{sec:results}

This section evaluates the proposed methods on two benchmarks. The \texttt{a280}-derived instances (Section~\ref{sec:a280-study}) retain the geometry and features of the classical TTP configuration and impose no limit on sortie length. In these instances, all methods from Section~\ref{sec:methodology} are compared, with the exact solver providing the reference solution wherever it can be applied. The \texttt{ttd300} benchmark (Section~\ref{sec:ttd300}), introduced in this paper, is a synthetic testbed in which drone endurance is the only controlled variable; it carries forward the strongest method from each family identified on \texttt{a280}, namely SA, GAT, and LISA. Throughout, every reported solution is recomputed with the same objective function under the exact load–speed law, independently of the method that produced it; $E_{\mathrm{D}}$ is additionally enforced on \texttt{ttd300}. The source code, datasets, and trained policy checkpoints used in this study are publicly accessible at \url{https://github.com/corbit-lab/ttpd}.

\paragraph{Compute environment}
All experiments were run under identical conditions on the same hardware: an Intel Xeon Platinum 8581C ($8$ vCPUs, $62$ GB RAM) running Debian GNU/Linux~12. The MILP was solved with Gurobi Optimizer 13.0 in Python 3.14.3 under a $24$-hour runtime limit and a $0.01\%$ relative MIP-gap target per instance. The five methods evaluated with multiple random seeds (VNS, SA, MLP, GAT, and LISA) were all executed on this host, so the reported objective values and runtimes are directly comparable across methods. 
Policy training was performed separately on an NVIDIA RTX 4000 Ada Generation GPU equipped with 20 GB VRAM, 50 GB RAM, and 9 vCPUs. For each benchmark and problem size, a separate policy is trained on instances drawn from the corresponding test distribution. This applies to the GAT policy, the MLP ablation, and the cloned policy within LISA. All reported measurements, including policy inference, were performed on the CPU host.

\subsection{Experiments on the a280 Benchmark}
\label{sec:a280-study}

The full $280$-city instance of the \texttt{a280} benchmark, drawn from the TTP suite of \citet{polyakovskiy2014comprehensive}, lies far beyond the practical reach of an exact solver. Customers are therefore sampled from its cities to construct smaller instances that preserve the spatial structure and item characteristics of the original while remaining computationally tractable for comparison with the MILP solver.

\subsubsection{Dataset and Environment}
\label{sec:dataset-env}

Table~\ref{tab:a280_params} lists the benchmark parameters. The renting ratio $R = 72.70$ is the dominant parameter, penalising every unit of mission time in the objective function~\eqref{eq:obj}. At the sizes considered, that charge exceeds the maximum profit the fleet can collect, so $G$ is negative on every benchmark instance; since the objective is maximised, less negative values are better. Performance is therefore driven by mission time rather than collected profit, favouring short truck routes and sorties that keep weight off the truck.

\begin{table}[h]
\centering
\caption{Parameters of the \texttt{a280}-derived TTP-D benchmark.}
\label{tab:a280_params}
\small
\begin{tabular}{ll}
\toprule
\textbf{Parameter} & \textbf{Value} \\
\midrule
Instance name & a280-TTP \\
Dimensions & 280 nodes, 1395 items \\
Renting ratio ($R$) & 72.70 \\
Knapsack capacity & $637{,}010$ (scaled per size to $W = 637{,}010\,N/280$) \\
Truck speed range & $v \in [0.1, 1.0]$ \\
Drone speed factor & $2.0 \times v_{\max}$ \\
\bottomrule
\end{tabular}
\end{table}

Since the TTP instances describe only truck movement, each sampled instance is extended with the components the TTP-D requires. The drone speed is set to $2.0\,v_{\max}$, and its payload limit is set to the weight of the heaviest item in the sampled instance, so that every item is eligible for drone collection. Sortie length is unbounded here, so the drone may fly arbitrarily far between launch and rendezvous; we drop this assumption in Section~\ref{sec:ttd300} and enforce real-world endurance constraints for the drone.

\paragraph{Benchmark instances and evaluation protocol}
For every size $N \in \{5, 10, 15, 20, 30, 40, 50\}$ we generate five benchmark instances, each a distinct subset of customers drawn from the $279$ non-depot cities of the \texttt{a280} benchmark. The depot is fixed at node~$1$, and every node contributes its single highest-profit item. All methods are scored on the same five draws at a given size. For consistency, each learning-based and heuristic method (GAT, MLP, SA, and VNS) is run over ten independent random seeds on each of the five instances, and every reported value is the mean across those runs. The deterministic MILP is solved once per instance. The metaheuristics run to fixed per-size budgets $B(N) = \{10, 30, 120, 300, 450, 600, 750\}$ seconds for $N = \{5, 10, 15, 20, 30, 40, 50\}$, respectively.

\subsubsection{Exact Solver Results}
\label{sec:exact_solver_results}
The MILP formulation was solved on five benchmark instances for each problem size $N \in \{5, 10, 15, 20\}$  from Section~\ref{sec:dataset-env}, with the weight-dependent truck velocity linearised using $K=10$ breakpoints. Beyond $N = 20$, the model became computationally intractable within the allocated time limit; exact results are therefore reported only for these four instance sizes.

\paragraph{Solver settings}
Gurobi is configured with an explicit emphasis on proving the optimality bound (\texttt{MIPFocus} $=3$), aggressive presolve (\texttt{Presolve} $=2$) and cut generation (\texttt{Cuts} $=2$), and a 20\% runtime budget allocation for internal heuristics (\texttt{Heuristics} $=0.2$). Each solve is warm-started from a greedy nearest-neighbour truck tour with a knapsack-feasible packing, and runs that do not certify the gap target within the runtime limit return the best incumbent.

\begin{figure}[htbp]
\centering
\includegraphics[width=0.75\textwidth]{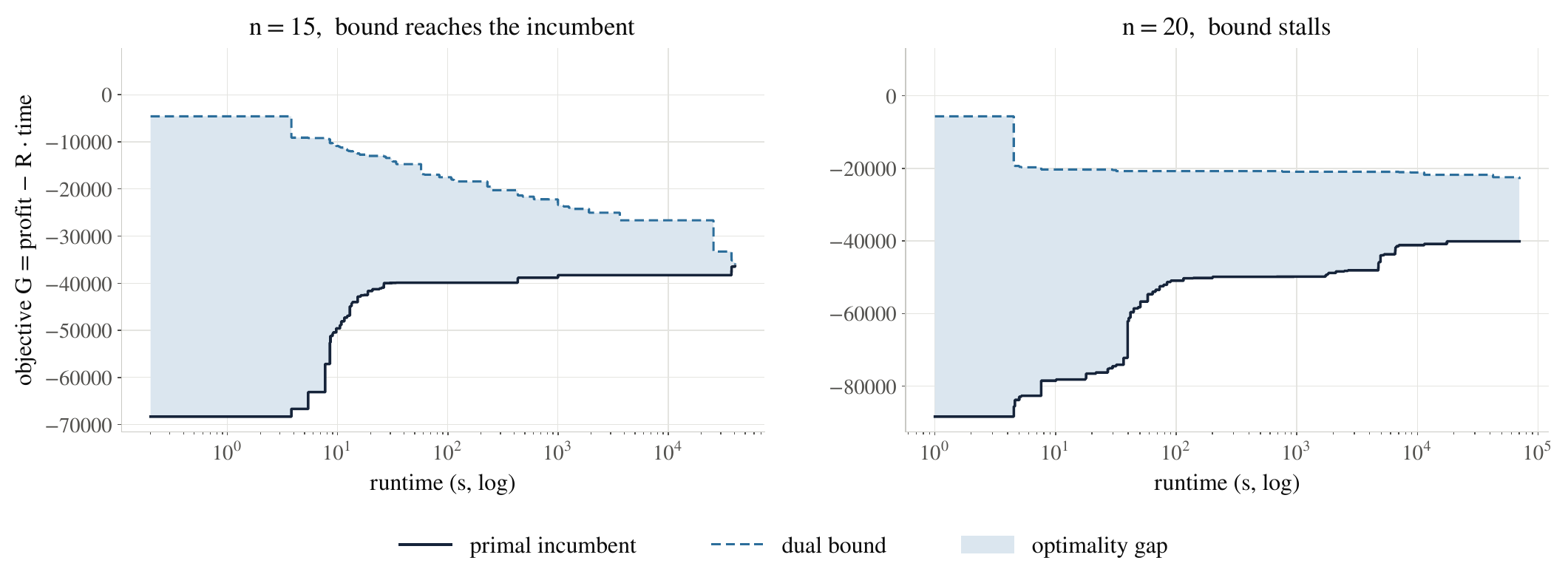}
\caption{MILP solver convergence on a benchmark instance for each of the two largest solvable problem sizes. The plot shows the evolution of the primal incumbent, the dual bound, and the resulting optimality gap (shaded area).}
\label{fig:milp_convergence}
\end{figure}

The exact solver solves every instance to proven optimality for $N = 5$ and $N = 10$, with mean total runtimes of $0.5$ and $49.1$ seconds, respectively, where the total comprises model-build and solve times.
At $N = 15$, two of the five instances are certified optimal and the remaining three exhaust the $24$-hour runtime limit, leaving a mean relative gap of $5.94\%$. At $N = 20$, all five instances reach the runtime limit with an open gap, averaging $40.05\%$. 
Table~\ref{tab:results} reports these incumbents alongside the other methods as a performance gap, the mean shortfall (\%) relative to the overall best-known solution (BKS) across all evaluated methods, which we reserve for comparisons against. 
Figure~\ref{fig:milp_convergence} illustrates the reduction of the optimality gap for two benchmark instances during the execution of the MILP solver.

Across each instance class, the resulting solution offloads a substantial proportion of customer visits to the drone, accounting for $3/5$ at $N = 5$, $4/10$ at $N = 10$, $7/15$ at $N = 15$, and $10/20$ at $N = 20$. This strategic offloading maintains a lighter payload on the ground vehicle, thereby preserving higher travel speeds. Figure~\ref{fig:milp_routes} illustrates the resulting structure for $N = 10$ and $N = 15$. Since every customer location is depicted, open circles explicitly identify customers who are visited but whose items are not collected, a distinction that a purely route-focused interpretation would overlook. All ten items are collected at $N=10$. However, at $N=15$, four customers ($4/15$) have items that remain uncollected, comprising two along the truck route and two visited via empty drone sorties. 


\subsubsection{Method Comparison}
\label{sec:a280-comparison}

Since the exact solver certifies optimality only for the smallest instances, solution quality is reported relative to the BKS, the best objective attained on an instance by any method across all runs. Performance gaps are computed as the mean over the five instances of $(\,\mathrm{BKS} - G\,)/|\mathrm{BKS}|$ for the seed-averaged objective $G$. Table~\ref{tab:results} reports, for each method and size, the performance gap (\%) and the mean runtime per instance in seconds. The last row gives the mean BKS objective per size, and the final column averages each method's gap across the sizes it covers. A smaller gap is better, and the smallest gap in each column is highlighted in bold. The two learned policies are GAT and a per-node MLP, both trained with PPO, as described in Section~\ref{sec:drl}. The table also includes the LISA at budget fractions $\beta \in \{5, 10, 20, 25, 33.3, 50\}\%$ of the annealing budget, on the same five instances.

\begin{table}[h]
\centering
\caption{Performance gap (\%) and mean runtime per instance (seconds) on the \texttt{a280} TTP-D benchmark, per instance size. The results for LISA are categorised by the budget fraction $\beta$, with its runtime dependent on the corresponding budget $\beta B(N)$.}

\label{tab:results}
\renewcommand{\arraystretch}{1.2}
\setlength{\tabcolsep}{3.5pt}
\footnotesize
\resizebox{\tabfit}{!}{%
\begin{tabular}{@{}l rr rr rr rr rr rr rr r@{}}
\toprule
\multirow{2}{*}{Method} & \multicolumn{2}{c}{$N=5$} & \multicolumn{2}{c}{$N=10$} & \multicolumn{2}{c}{$N=15$} & \multicolumn{2}{c}{$N=20$} & \multicolumn{2}{c}{$N=30$} & \multicolumn{2}{c}{$N=40$} & \multicolumn{2}{c}{$N=50$} & \multirow{2}{*}{Mean} \\
\cmidrule(lr){2-3}\cmidrule(lr){4-5}\cmidrule(lr){6-7}\cmidrule(lr){8-9}\cmidrule(lr){10-11}\cmidrule(lr){12-13}\cmidrule(lr){14-15}
 & Gap & Time & Gap & Time & Gap & Time & Gap & Time & Gap & Time & Gap & Time & Gap & Time & \\
\midrule
MILP & \textbf{0.00} & 0.5 & \textbf{0.00} & 49.1 & 0.36 & 66{,}945 & 2.61 & 86{,}400 & \multicolumn{2}{c}{$-$} & \multicolumn{2}{c}{$-$} & \multicolumn{2}{c}{$-$} & {$-$} \\
SA & \textbf{0.00} & 10 & 0.22 & 30 & \textbf{0.00} & 120 & \textbf{0.09} & 301 & \textbf{0.00} & 451 & \textbf{0.00} & 600 & \textbf{0.00} & 750 & \textbf{0.04} \\
VNS & \textbf{0.00} & 10 & 0.22 & 30 & 0.32 & 120 & 0.19 & 300 & 5.14 & 450 & 17.28 & 600 & 27.07 & 750 & 7.17 \\
MLP & \textbf{0.00} & 3.6 & 4.85 & 9.6 & 3.88 & 25 & 5.41 & 43 & 15.09 & 25.8 & 21.30 & 58.6 & 19.12 & 40.6 & 9.95 \\
GAT & 0.34 & 3.5 & 2.69 & 11 & 3.91 & 25 & 4.91 & 40 & 11.48 & 23.1 & 15.68 & 50.3 & 19.92 & 41.6 & 8.42 \\
\midrule
LISA, $\beta = 5\%$ & \textbf{0.00} & 0.5 & 0.88 & 1.5 & 2.20 & 6 & 2.76 & 15 & 6.16 & 22.5 & 10.60 & 30 & 12.94 & 37.5 & 5.08 \\
LISA, $\beta = 10\%$ & \textbf{0.00} & 1 & 0.53 & 3 & 1.23 & 12 & 1.68 & 30 & 4.98 & 45 & 7.76 & 60 & 10.92 & 75 & 3.87 \\
LISA, $\beta = 20\%$ & \textbf{0.00} & 2 & 0.49 & 6 & 1.23 & 24 & 1.38 & 60 & 3.06 & 90 & 5.89 & 120 & 8.19 & 150 & 2.89 \\
LISA, $\beta = 25\%$ & \textbf{0.00} & 2.5 & 0.49 & 7.5 & 0.53 & 30 & 1.38 & 75 & 2.96 & 112.5 & 5.44 & 150 & 8.19 & 187.5 & 2.71 \\
LISA, $\beta = 33.3\%$ & \textbf{0.00} & 3.3 & 0.49 & 10 & 0.11 & 40 & 1.38 & 100 & 2.31 & 150 & 5.76 & 200 & 6.84 & 250 & 2.41 \\
LISA, $\beta = 50\%$ & \textbf{0.00} & 5 & 0.02 & 15 & 0.11 & 60 & 1.37 & 150 & 2.31 & 225 & 1.82 & 300 & 5.53 & 375 & 1.59 \\
\midrule
BKS & \multicolumn{2}{c}{$-30{,}619$} & \multicolumn{2}{c}{$-33{,}698$} & \multicolumn{2}{c}{$-36{,}739$} & \multicolumn{2}{c}{$-37{,}888$} & \multicolumn{2}{c}{$-37{,}710$} & \multicolumn{2}{c}{$-37{,}311$} & \multicolumn{2}{c}{$-36{,}965$} & \\
\bottomrule
\end{tabular}%
}
\end{table}

\paragraph{Small instances} For $N = 5$ instances, MILP, SA, VNS, and MLP consistently reach the BKS, whereas the GAT policy trails slightly by $0.34\%$, a shortfall driven entirely by a single instance. At $N = 10$, the MILP solver achieves optimal solutions across all instances, while SA and VNS match the BKS on four instances and lag by $1.08\%$ on the fifth, yielding a mean gap of $0.22\%$. The learned neural policies fall further behind, with GAT at $2.69\%$ and MLP at $4.85\%$. Meanwhile, LISA matches the BKS across all budget fractions at $N = 5$; at $N = 10$, its half-budget variant ($\beta = 50\%$) achieves a $0.02\%$ gap in just 15~seconds, outperforming both metaheuristics in half the runtime.

\paragraph{Mid-sized instances} Relative performance shifts for mid-sized instances. At $N = 15$, SA attains the BKS on all instances, followed closely by MILP ($0.36\%$) and VNS ($0.32\%$), while both learned policies remain near $3.9\%$. The slight non-zero gap for MILP is attributable to the $K = 10$ piecewise-linear approximation: the solver optimises a surrogate model, whereas all final solutions are evaluated against the exact load-speed law. At $N = 20$, metaheuristics dominate: SA ($0.09\%$) and VNS ($0.19\%$) substantially outperform the MILP incumbent ($2.61\%$). Here, the gap to the exact solver is no longer an approximation artefact; after 24 hours, the solver terminates with a large open optimality gap of $\approx 40\%$, whereas five minutes of simulated annealing achieves superior objective values. Across both sizes, LISA bridges the gap between metaheuristics and learned policies, reaching $0.11\%$ and $1.38\%$, respectively, with a computational budget of $\beta \ge 33.3\%$.

\paragraph{Large Instances} For large instances with $N \ge 30$, SA establishes all BKS solutions, fixing its gap at $0.00\%$ by construction. 
In contrast, VNS performance deteriorates significantly as problem scale increases, rising from $5.14\%$ at $N = 30$ to $27.07\%$ at $N = 50$, as a single shake-and-descent trajectory inadequately covers the expanding search space within the fixed time budget. 
Overall, SA achieves the lowest mean gap (0.04\%), followed by VNS (7.17\%), GAT (8.42\%), and MLP (9.95\%). Meanwhile, LISA at $\beta = 50\%$ maintains strong solution quality across large instances with mean gaps of 2.31\%, 1.82\%, and 5.53\%, outperforming both VNS and the learned policies while requiring significantly less computation than a full metaheuristic search. 

Run-to-run variability does not account for the relative performance ranking of the two metaheuristics. Starting at $N = 40$, the gap between the mean performances of SA and VNS in Table~\ref{tab:results} is strictly greater than the sum of their seed-to-seed standard deviations, confirming that the ranking at larger instance scales is statistically robust rather than an artefact of seed selection. For smaller problem sizes, the two metaheuristics should be interpreted as statistically equivalent. The dispersion details are available in Table~S3 of the Supplementary Material.

The runtimes presented in Table~\ref{tab:results} exhibit an inverse relationship with solution quality. Specifically, the exact solver is computationally viable only for the two smallest instance sizes, whereas the learned policies complete decoding in under a minute. Furthermore, the comparison against the per-node MLP explicitly isolates the impact of the encoder architecture: both models utilise an identical decoder, masking mechanism, and training pipeline while achieving equivalent inference speeds. Consequently, the superior solution quality of the GAT model incurs zero additional inference overhead.

\subsection{Experiments on the ttd300 Benchmark}
\label{sec:ttd300}

Our experiments on \texttt{a280}-derived instances sample customer locations and item characteristics from a fixed list while assuming unbounded drone endurance: the drone may fly arbitrarily far between launch and rendezvous.
In practice, battery capacity restricts the round-trip sortie and is often the primary operational bottleneck, superseding payload or speed as the limiting resource. To isolate this factor, we introduce \texttt{ttd300}, a synthetic TTP-D benchmark comprising randomly generated instances that feature drone endurance parameters. We use this benchmark to evaluate the strongest representative of each method family identified in Section~\ref{sec:a280-comparison}: SA, GAT, and the LISA method.

\subsubsection{Benchmark Design}
\label{sec:ttd300-bench}

Customer locations are drawn uniformly at random as integer coordinates within a fixed $[0, 300]^2$ bounding box, with distances calculated as the ceiling of the Euclidean distance. This uniform spatial distribution deliberately departs from the previous evaluation dataset; for instance, the \texttt{a280} layout is highly non-uniform and fails a chi-squared uniformity test. In contrast, a uniform, fixed-size domain ensures that the endurance radius retains a consistent physical interpretation across all instances. The benchmark spans seven problem sizes, $N \in \{10, 20, 30, 40, 50, 75, 100\}$, each containing five independent instances or layouts (L1-L5). Each location offers five items drawn from the \texttt{a280} ``uncorrelated, similar weights'' class ($w \sim U[1000, 1009]$ and $p \sim U[1, 1000]$). Following the existing experimental protocol, the baseline models collect at most the single most profitable item per city; however, the complete item lists are utilised in the multi-item study detailed in Section~\ref{sec:ttd-multi-item}. The knapsack capacity maintains the previous per-city rate, fixed at $W = \lfloor 2275.0357\,N \rceil$ across all layouts for a given size, where $\lfloor x \rceil$ denotes $x$ rounded to the nearest integer. Finally, while the truck speed range and drone speed factor mirror the earlier setup, the rental ratio is reduced to $R = 50$. Table~\ref{tab:ttd300_params} summarises the parameter values for the \texttt{ttd300} instances.

The drone endurance $E_{\mathrm{D}}$ restricts the maximum round-trip distance of a sortie: a flight launched at $\ell$, visiting $k$, and rejoining at $j$ is admissible only if $d_{\ell k} + d_{kj} \leq E_{\mathrm{D}}$. Each layout is generated with four \emph{relative} endurance fractions, $E_{\mathrm{D}} = f \cdot d_{\max}$ for $f \in \{0.25, 0.5, 0.75, 1.0\}$, where $d_{\max}$ is the maximum pairwise distance within that specific instance. This yields a total of 140 instances. This limit is strictly enforced across all solvers: the metaheuristic evaluator rejects any plan containing an excessively long sortie, while the DRL policy’s launch and rejoin masks actively exclude endurance-infeasible actions. Furthermore, the DRL framework explicitly captures endurance by deriving the normalised endurance, $E_{\mathrm{D}}/2d_{\max}$, and the in-flight slack, $E_{\mathrm{D}} - (d_{\ell k} + d_{k c_t})$.

\begin{table}[h]
\centering
\caption{Parameters of the \texttt{ttd300} TTP-D benchmark.}
\label{tab:ttd300_params}
\small
\begin{tabular}{ll}
\toprule
\textbf{Parameter} & \textbf{Value} \\
\midrule
Topology & Uniform integer coordinates on $[0, 300]^2$, ceiling of Euclidean distance \\
Sizes & $N \in \{10, 20, 30, 40, 50, 75, 100\}$, 5 layouts (L1--L5) each \\
Items & $5$ per customer, $w \sim U[1000, 1009]$, $p \sim U[1, 1000]$ \\
Capacity & $W = \lfloor 2275.0357\,N \rceil$, fixed per size \\
Rental ratio ($R$) & $50.0$ \\
Truck speed range & $v \in [0.1, 1.0]$ \\
Drone speed factor & $2.0 \times v_{\max}$ \\
Endurance & $E_{\mathrm{D}} = \lfloor f \cdot d_{\max} \rceil$ for $f \in \{0.25, 0.5, 0.75, 1.0\}$ \\
\bottomrule
\end{tabular}
\end{table}

The SA algorithm is executed within fixed, size-dependent runtime budgets $B(N) = \{120, 300, 450, 600, 750, 975, 1200\}$ seconds for $N = \{10, 20, 30, 40, 50, 75, 100\}$, respectively, using five independent seeds per instance. The GAT policy is retrained for each size on the \texttt{ttd300} instance distribution, with the endurance fraction randomly drawn per episode. It uses the PPO and POMO configurations and decodes each instance via a single beam search of width 256 across eight dihedral views. Given the absence of an exact solution reference at these scales, performance gaps are measured against a per-instance reference: the best validated solution found by the SA solver across all seeds using the full budget $B(N)$.

\subsubsection{Results}
\label{sec:ttd300-results}

\begin{figure}[htbp]
\centering
\begin{subfigure}[t]{0.47\textwidth}
  \centering
  \includegraphics[width=0.94\textwidth]{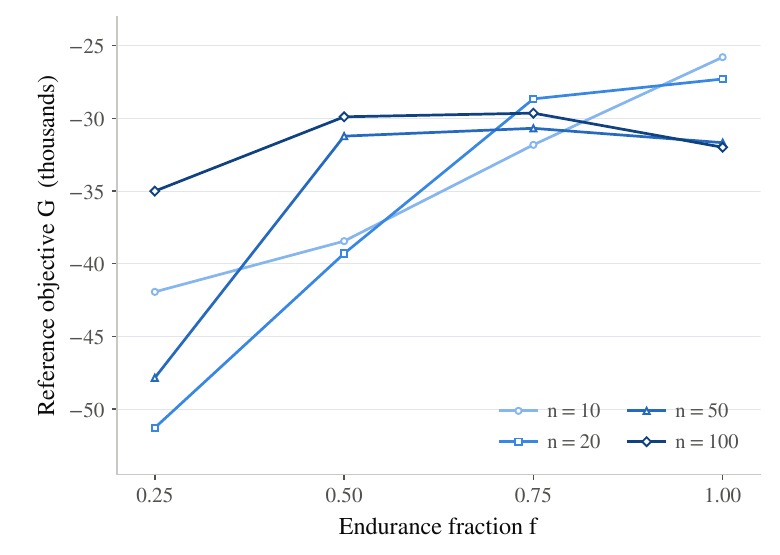}
  \caption{Reference objective against the endurance fraction.}
  \label{fig:sa_endurance-a}
\end{subfigure}
\hfill
\begin{subfigure}[t]{0.47\textwidth}
  \centering
  \includegraphics[width=0.94\textwidth]{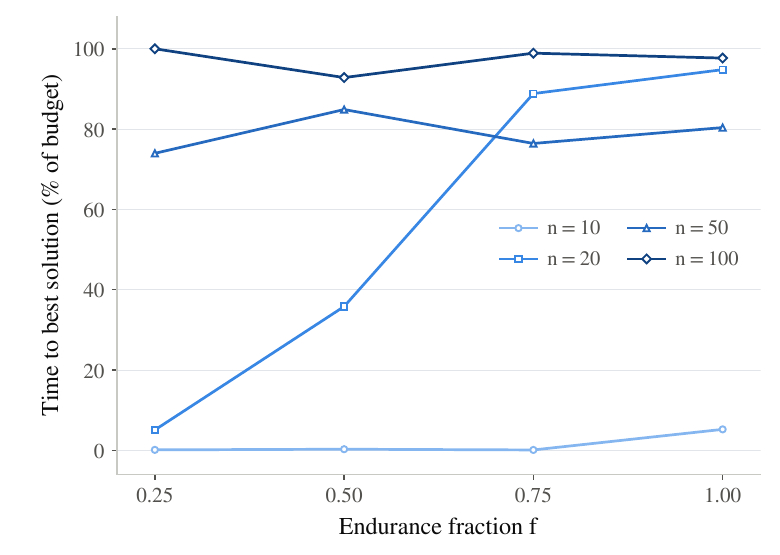}
  \caption{Fraction of budget $B(N)$ at last search improvement.}
  \label{fig:sa_endurance-b}
\end{subfigure}
\caption{The full-budget SA reference across the four endurance fractions, on instance sizes of \texttt{ttd300}. Each point is the mean over the five layouts at that setting.}
\label{fig:sa_endurance}
\end{figure}

We report the results in two steps: first, along the endurance axis that the benchmark was designed to isolate, and second, across the evaluated methods. Since no method certifies optimality on \texttt{ttd300}, the endurance axis is based on the reference solutions rather than on the optimality gap of any individual method. Table~S8 of the Supplementary Material reports the full-budget SA reference at each $(N, f)$ setting, giving the net objective $G$, the makespan $\tau_{\depotend}$ that the rental term charges for, and the point in the budget at which the search last improved. 
Figure~\ref{fig:sa_endurance} illustrates these findings. Since this SA reference defines the baseline gap in Table~\ref{tab:ttd300_results}, its own gap is zero by construction. A plan feasible at $f$ stays feasible at any $f' > f$, so the attainable objective is non-decreasing in $f$, and any decrease is a property of the search. However, as shown in the figure, the graph is not strictly growing; this discrepancy arises from the heuristic nature of the search methods.

\paragraph{The endurance axis} Loosening the sortie radius from $f = 0.25$ to $f = 0.50$ improves the objective at every size and across all $35$ layouts, by $8.3\%$ of $|G|$ at $N = 10$, rising to $34.7\%$ at $N = 50$. Beyond $f = 0.50$, the curve flattens, and two regimes emerge. At $N \leq 20$, the objective keeps improving through $f = 1.00$, because the tight radius effectively suppresses sortie deployment: at $N = 10$ and $f = 0.25$, the reference solution launches no sortie in four of the five layouts. From $N = 30$ upward, the first step accounts for most of the improvement, while the three remaining settings differ by at most $9.8$ points. The gain comes from makespan, since at $N \geq 30$ the fleet already collects almost every item at the tightest radius. At $N = 50$, the first step is worth $16{,}622$, of which $14{,}728$ comes from the rental term as the makespan falls from $1{,}740$ to $1{,}445$. With a short radius, the truck must detour to customers the drone cannot reach, and the rental clock charges for every detour. 

Runtime is the same across settings because SA runs to the budget $B(N)$ in every case, so endurance only changes how much of that budget the search uses. At $N = 20$, the last improvement moves from $5\%$ of the budget at $f = 0.25$ to $95\%$ at $f = 1.00$, as a wider radius admits more launch and rendezvous pairings and enlarges the neighbourhood to be covered. At $N = 75$ and $N = 100$, it arrives after $88\%$ of the budget at every setting, suggesting that the reference is still limited by the budget rather than converged. That is the most likely source of the small reversals at $N \geq 50$, where the objective declines by $450$ to $2{,}096$ between $f = 0.50$ and $f = 1.00$ even though the feasible set grows. Section~\ref{sec:ttd-drone-speed} examines the same axis in the drone speed sweep, where the same saturation limits the benefit of a faster drone.

Table~\ref{tab:ttd300_results} compares the methods by reporting the gap to the reference and the mean runtime per instance for each size. The final column averages each method's gap across all seven sizes. The GAT results represent the average of the beam decode at each size. The table also includes the LISA evaluated across a ladder of budget fractions $\beta \in \{5, 10, 20, 33.3, 50\}\%$ of the SA budget, evaluated against the same reference. Section~\ref{sec:ttd300-pareto} further analyses these results from the perspective of solution quality versus runtime budget.

\begin{table}[h]
\centering
\caption{Performance gap (\%) to the SA metaheuristic reference and mean runtime per instance (seconds) on the \texttt{ttd300} benchmark.}

\label{tab:ttd300_results}
\renewcommand{\arraystretch}{1.2}
\setlength{\tabcolsep}{3.5pt}
\footnotesize
\resizebox{\tabfit}{!}{%
\begin{tabular}{@{}l rr rr rr rr rr rr rr r@{}}
\toprule
\multirow{2}{*}{Method} & \multicolumn{2}{c}{$N=10$} & \multicolumn{2}{c}{$N=20$} & \multicolumn{2}{c}{$N=30$} & \multicolumn{2}{c}{$N=40$} & \multicolumn{2}{c}{$N=50$} & \multicolumn{2}{c}{$N=75$} & \multicolumn{2}{c}{$N=100$} & \multirow{2}{*}{Mean} \\
\cmidrule(lr){2-3}\cmidrule(lr){4-5}\cmidrule(lr){6-7}\cmidrule(lr){8-9}\cmidrule(lr){10-11}\cmidrule(lr){12-13}\cmidrule(lr){14-15}
 & Gap & Time & Gap & Time & Gap & Time & Gap & Time & Gap & Time & Gap & Time & Gap & Time & \\
\midrule
SA (reference) & 0.00 & 120 & 0.00 & 300 & 0.00 & 450 & 0.00 & 600 & 0.00 & 750 & 0.00 & 975 & 0.00 & 1{,}200 & 0.00 \\
GAT & 2.37 & 25 & 18.04 & 72 & 29.77 & 137 & 26.05 & 260 & 37.84 & 458 & 33.26 & 1{,}372 & 51.05 & 3{,}098 & 28.34 \\
\midrule
LISA, $\beta = 5\%$ & 0.00 & 6 & 0.85 & 15 & 5.81 & 22.5 & 5.23 & 30 & 13.60 & 37.5 & 17.82 & 48.8 & 31.65 & 60 & 10.71 \\
LISA, $\beta = 10\%$ & 0.00 & 12 & 0.76 & 30 & 2.64 & 45 & 6.51 & 60 & 10.83 & 75 & 15.33 & 97.5 & 18.00 & 120 & 7.72 \\
LISA, $\beta = 20\%$ & 0.00 & 24 & 0.33 & 60 & 2.95 & 90 & 4.09 & 120 & 8.38 & 150 & 11.80 & 195 & 17.62 & 240 & 6.45 \\
LISA, $\beta = 33.3\%$ & 0.00 & 40 & 0.15 & 100 & 2.94 & 150 & 3.67 & 200 & 4.42 & 250 & 10.28 & 325 & 17.18 & 400 & 5.52 \\
LISA, $\beta = 50\%$ & 0.00 & 60 & 0.20 & 150 & 2.94 & 225 & 3.39 & 300 & 4.42 & 375 & 9.56 & 488 & 14.93 & 600 & 5.06 \\
\midrule
Reference $G$ & \multicolumn{2}{c}{$-34{,}497$} & \multicolumn{2}{c}{$-36{,}632$} & \multicolumn{2}{c}{$-34{,}930$} & \multicolumn{2}{c}{$-36{,}410$} & \multicolumn{2}{c}{$-35{,}354$} & \multicolumn{2}{c}{$-31{,}823$} & \multicolumn{2}{c}{$-31{,}629$} & \\
\bottomrule
\end{tabular}%
}
\end{table}

SA anchors the benchmark in Table~\ref{tab:ttd300_results}, with the best full-budget SA solution serving as the reference for each instance. The GAT policy trails consistently throughout, despite being retrained per size on the new data distribution, recording a gap of $2.37\%$ at $N = 10$, roughly $18$-$38\%$ at intermediate sizes, and $51.05\%$ at $N = 100$, demonstrating that a constructive single pass cannot compete with search-based methods in solution quality on this benchmark. This poor performance can also be attributed to shifts in problem features, specifically drone endurance, as well as the fact that the hyperparameters were optimised for the \texttt{a280} benchmark dataset. Nevertheless, its primary advantages remain its speed at small to intermediate sizes and its utility as a warm-start generator within the LISA framework. Furthermore, the SA reference is sufficiently stable to support these conclusions; across the five seeds, the per-instance standard deviation of the full-budget SA objective averages between $1.9\%$ and $5.5\%$ for sizes $N = 30$ to $N = 100$.

\begin{figure}[htbp]
\centering
\includegraphics[width=0.75\textwidth]{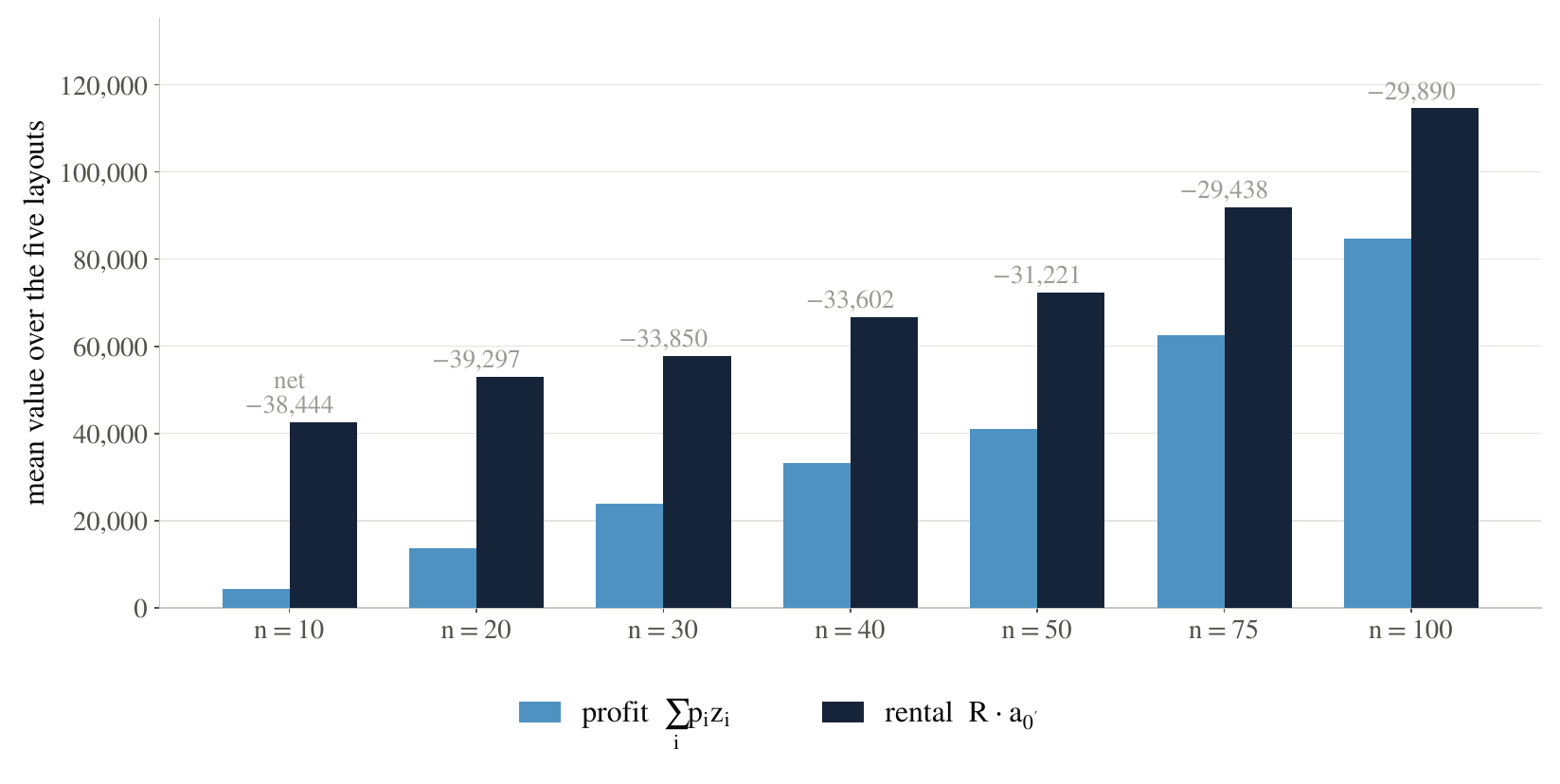}
\caption{Profit against rental cost at the \texttt{ttd300} SA reference, where endurance fraction $f = 0.5$. Each bar is the mean over the five layouts at that size, annotated with the net objective $G$.}
\label{fig:obj_decomp}
\end{figure}

Figure~\ref{fig:obj_decomp} decomposes this set of the best objective values into its profit and rental components at a selected endurance fraction $f=0.50$, whose rental term is $R\,\tau_{\depotend}$; each bar represents the best validated SA solution per instance across all seeds at the full budget $B(N)$. At $R = 50$, the rental term exceeds the collected profit at every size, meaning no operating point is purely profitable. However, this shortfall narrows as $N$ grows, because profit accumulates faster than the mission duration lengthens. Specifically, the profit-to-rental ratio climbs from $0.10$ at $N = 10$ to $0.74$ at $N = 100$, narrowing the net loss from $-38{,}444$ to $-29{,}890$. Aggregating these results across all four endurance fractions yields the reference row shown in Table~\ref{tab:ttd300_results} for every size.

\paragraph{Runtime} 
The single-pass DRL policy is computationally cheaper up to $N = 50$. Beyond this, the beam search across eight dihedral views scales superlinearly on the CPU host, overtaking the full annealing budget at the two largest instance sizes. Evaluating the methods at $N = 40$, a size shared by both benchmarks, clarifies the runtime-quality trade-offs. On \texttt{a280}, LISA at $\beta = 50\%$ closes the gap to $1.82\%$, effectively halving the annealing runtime to achieve near-reference quality. Conversely, standalone learned policies maintain gaps above $15\%$ despite incurring a computational cost comparable to LISA at $\beta = 10\%$. On \texttt{ttd300}, the LISA performance curve flattens at a $3.39\%$ gap for $\beta = 50\%$. Here, the single-pass policy matches the runtime of LISA at $\beta = 33.3\%$ but trails by $26.05\%$ in quality, as beam decoding scales worse than the annealing budget. Ultimately, as the exact MILP becomes intractable beyond $N = 20$, alternatives like VNS and standalone policies deteriorate significantly by $N = 50$, positioning LISA as a robust compromise between speed and quality.

\subsubsection{Budget Sensitivity of LISA}
\label{sec:ttd300-pareto}

\begin{figure}[htbp]
\centering
\begin{subfigure}[t]{0.45\textwidth}
  \centering
  \includegraphics[width=0.94\textwidth]{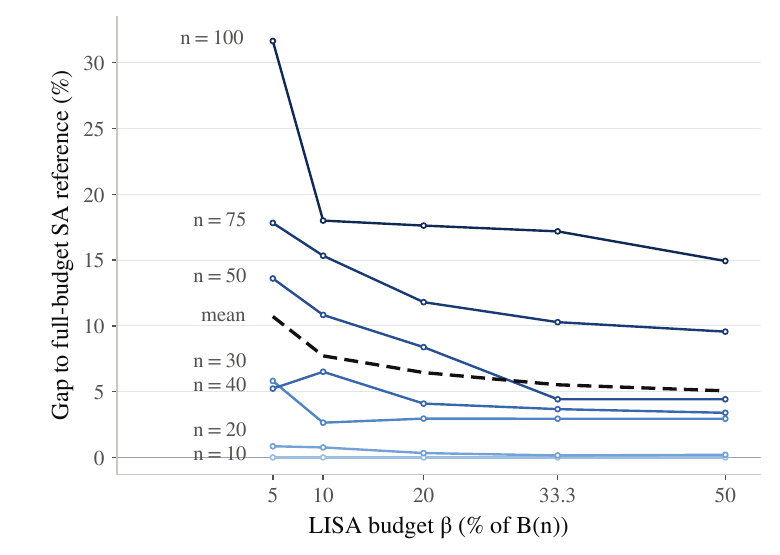}
  \caption {Gap against the budget fraction $\beta$, with one curve per instance size and the dashed mean over all 28 cells.}
  \label{fig:lisa-ttd300-a}
\end{subfigure}
\hfill
\begin{subfigure}[t]{0.45\textwidth}
  \centering
  \includegraphics[width=0.90\textwidth]{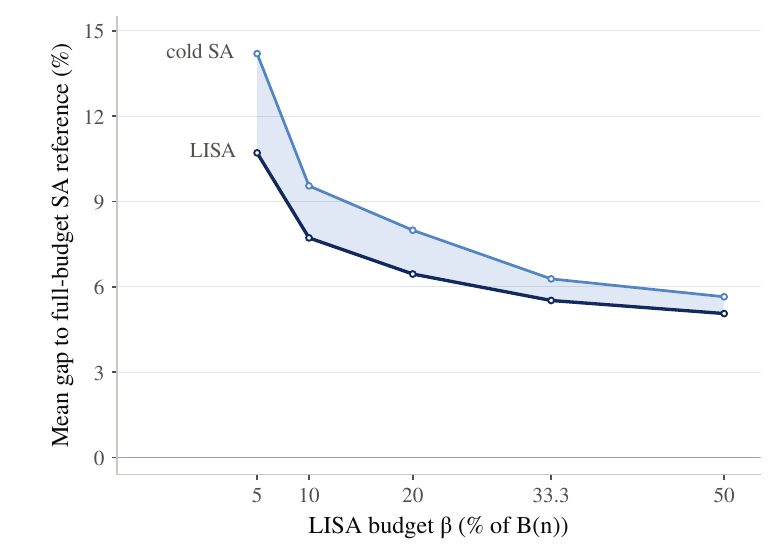}
  \caption {LISA against a cold annealing run at the same reduced budget, with the shaded band showing the warm-start gain.}
  \label{fig:lisa-ttd300-b}
\end{subfigure}
\caption{Performance of LISA on the \texttt{ttd300} L1 layouts.}
\label{fig:lisa-ttd300}
\end{figure}

The LISA architecture uses the budget fraction $\beta$ to interpolate between the cloned policy and the full metaheuristic. We evaluate this parameter on a fixed slice of the benchmark: the first layout (L1) across all seven instance sizes and all four endurance fractions. This yields 28 configurations, each solved as an independent run at $\beta \in \{5, 10, 20, 33.3, 50\}\%$ of the budget $B(N)$. Performance gaps are measured against the same reference as before: the best of the five full-budget SA seeds per instance. The LISA results in Table~\ref{tab:ttd300_results} report the mean gap across the four endurance fractions for each cell, the implied budget $\beta B(N)$ in the time column, and the overall mean; Figure~\ref{fig:lisa-ttd300} plots this profile.

In Figure~\ref{fig:lisa-ttd300-a}, the performance profile descends steeply at the low-budget end and then flattens. The cloned policy's unrepaired baseline averages a $64\%$ gap, but allocating just 5\% of the annealing budget reduces this to $10.71\%$. Subsequent budget increases yield diminishing returns, reaching a $5.06\%$ average gap at $\beta = 50\%$. While mean values decrease monotonically, individual stochastic runs occasionally fluctuate. LISA's value is most pronounced in small- to medium-sized instances: for sizes up to $N = 40$, it achieves within a $4\%$ gap of the reference using at most one-third of the full budget. Conversely, the largest instances retain gaps of roughly $10\%$ to $15\%$ even when given half the budget. Figure~\ref{fig:lisa-ttd300-b} isolates the contribution of the learned warm start, demonstrating that LISA consistently outperforms a standard, ``cold'' SA run given the identical reduced budget. At higher budgets, the annealing repair dominates and determines the final solution quality, with no consistent difference across endurance fractions.

\section{Sensitivity Analysis}
\label{sec:sensitivity}

The \texttt{ttd300} results reported thus far were obtained at a single baseline operating point: one collected item per city, a renting ratio $R = 50$, a baseline capacity $W = \lfloor 2275.0357\,N \rceil$, a minimum truck speed $v_{\min} = 0.1$, and a drone twice as fast as the truck ($\phi = 2$). This section explores deviations from this baseline along three axes: the viability of multi-item collection per city, the impact of drone speed on synchronisation and makespan, and the relative influence of key physical and economic parameters on the objective. 
The objective is linear with respect to $R$ through the single rental term $R\, \tau_{\depotend}$, whereas capacity and vehicle speeds influence the objective only via the nonlinear timing embedded within $\tau_{\depotend}$. Consequently, $R$ is expected to be the dominant driver of $G$, while physical parameters should exert a weaker, saturating effect.
Every configuration is evaluated across all four endurance fractions, illustrating how these conclusions shift as the sortie radius tightens. Due to computational challenges in evaluating all available instances, the sensitivity analysis is restricted to a single layout (L1) for each instance size. SA is employed to solve the instances because it consistently delivers superior solution quality, despite higher runtime. Unless stated otherwise, SA is executed once per instance using a fixed random seed. The sole exception is in Section~\ref{sec:ttd-multi-item}, where additional runs are conducted to effectively evaluate the significantly larger combinatorial search space of the multi-item configuration.

\subsection{Multi-item collection}
\label{sec:ttd-multi-item}

In the \texttt{ttd300} benchmark, the vehicle capacity was deliberately calibrated for a multi-item regime. The single most profitable item per city weighs approximately $1004$ on average; thus, the single-item protocol commits at most $\approx 1004\,N$ of weight against a capacity of $\approx 2275\,N$, meaning the knapsack constraint never binds. In contrast, the full five-item lists total $\approx 5022\,N$, forcing a capacity-constrained knapsack decision. Table~S4 in the Supplementary Material compares the two protocols at a common budget across sizes and endurance fractions, reporting the best result from five independent runs.

\begin{figure}[htbp]
\centering
\begin{subfigure}[t]{0.45\textwidth}
  \centering
  \includegraphics[width=0.90\textwidth]{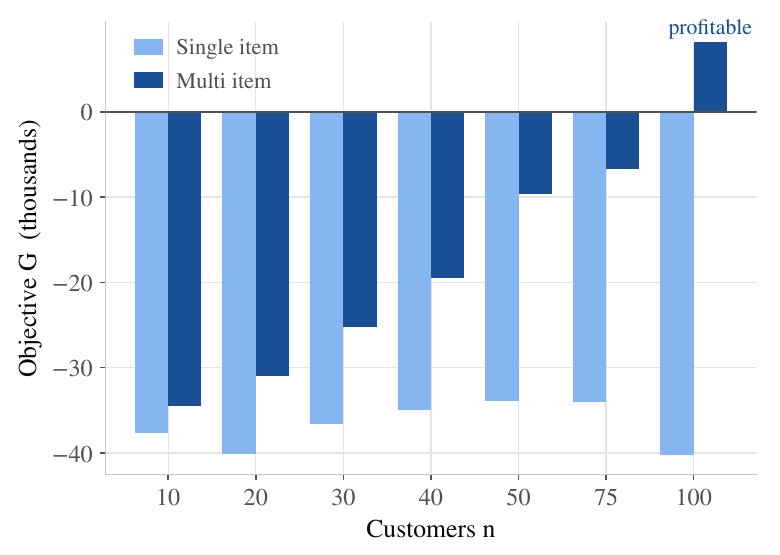}
  \caption{Objective per size on the L1 layouts.}
  \label{fig:multiitem-a}
\end{subfigure}
\hfill
\begin{subfigure}[t]{0.45\textwidth}
  \centering
  \includegraphics[width=0.90\textwidth]{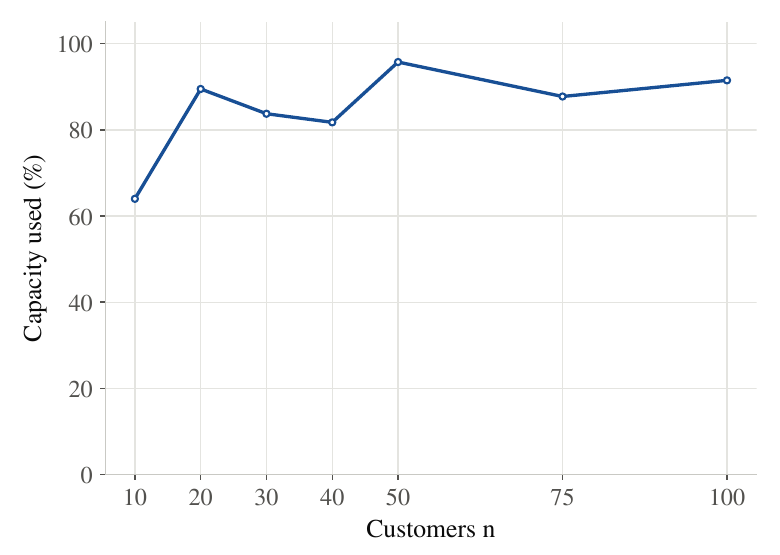}
  \caption{Multi-item collection knapsack utilisation.}
  \label{fig:multiitem-b}
\end{subfigure}
\caption{Comparison of single-item and multi-item collection on \texttt{ttd300}. The objective gains in (a) are directly driven by the higher capacity utilisation shown in (b).}
\label{fig:multiitem}
\end{figure}

Relaxing the single-item restriction is beneficial across every size and endurance fraction, with gains compounding at scale. The objective improvement ($\Delta G$) rises from between $1{,}626$ and $4{,}004$ at $N = 10$ to between $40{,}058$ and $53{,}945$ at $N = 100$. At the largest scale, the gain is substantial enough to yield a positive objective at every endurance fraction (peaking at $+13{,}342$ for $f = 1.00$), representing the only profitable operating points in our experiments. Rather than distributing collections evenly, the multi-item solution concentrates them, retrieving an average of $2.29$ to $3.25$ items per visited city and reaching the five-item limit at least once per size. Accordingly, capacity utilisation rises from roughly $44\%$ at $N = 10$ to $84$--$100\%$ for $N \geq 50$. The truck absorbs most of this additional mass; for instance, at $N = 100$, the truck retrieves items from $55$--$60$ cities compared to the drone's $21$--$26$. Across endurance fractions, the objective improves non-monotonically due to heuristic variance in larger instances. While the objective steadily increases with $f$ at $N = 10$, as a larger operating radius enables more retrievals, at $N = 100$ it fluctuates by roughly $14{,}000$ without a clear trend. Consequently, the single-item convention understates achievable profit. At the largest problem size, it is the difference between a net loss and a profitable operation. Figure~\ref{fig:multiitem} illustrates this comparison alongside the resulting knapsack utilisation.

\subsection{Drone speed}
\label{sec:ttd-drone-speed}

We sweep the drone speed factor $v_{\mathrm{D}} = \phi\, v_{\max}$ across $\phi \in \{0, 0.5, 1, 2, 3\}$, solving each $(\phi, N, f)$ configuration; $\phi = 0$ disables the drone and establishes a truck-only solution. This sweep interacts heavily with the endurance budget, as the radius dictates sortie feasibility. Table~S5 in the Supplementary Material reports the objectives and collection splits for every $(N, \phi)$ pair; the findings are summarised here.

\begin{figure}[htbp]
\centering
\includegraphics[width=0.7\textwidth]{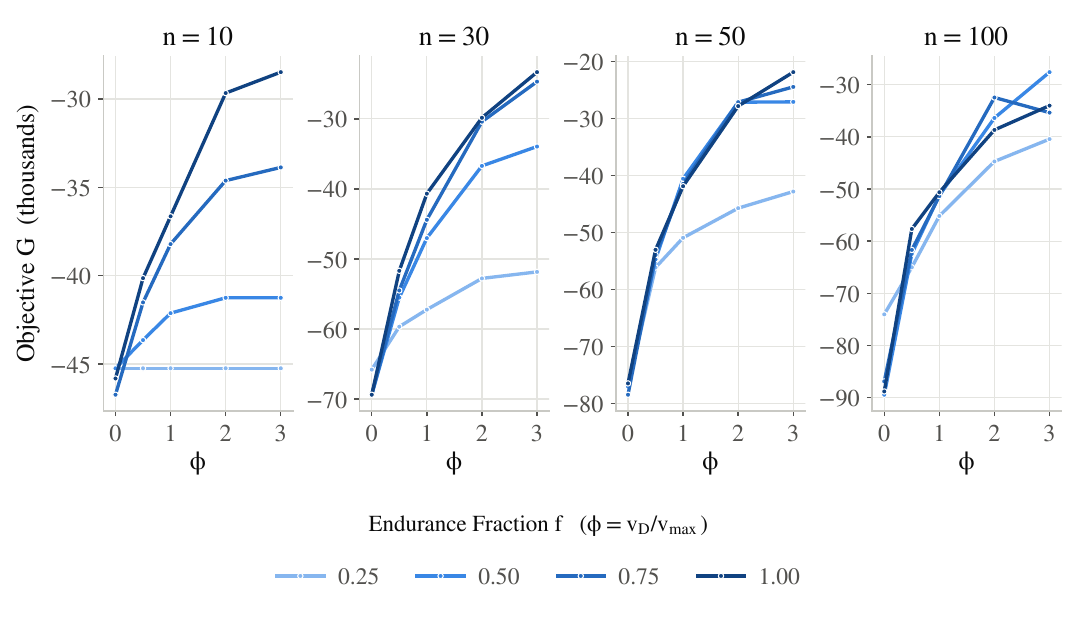}
\caption{Mean objective versus drone speed factor ($\phi$) across problem sizes. Initial speed increases yield the largest gains, followed by diminishing returns. Short endurance ($f=0.25$) severely restricts these gains, completely flattening the curve at $N=10$ where no sorties are feasible.}
\label{fig:speed_endurance}
\end{figure}

Figure~\ref{fig:speed_endurance} plots this experiment. The objective function is non-decreasing with respect to $\phi$ in nearly every $(N, f)$ block, with the most significant gains concentrating in the initial speed increments. For example, at $N = 100$ and $f = 0.50$, the truck-only baseline of $-89{,}453$ improves to $-62{,}420$ at half speed, $-51{,}200$ at $1\times$ truck speed, $-36{,}374$ at $\phi = 2$, and $-27{,}590$ at $\phi = 3$. Each successive speed increase yields diminishing returns. An exception occurs at $N = 100, f = 0.75$, where $\phi = 2$ outperforms $\phi = 3$ by approximately $2{,}900$. This is a consequence of the stochastic search budget at this scale, not a property of the problem itself.

Ultimately, drone endurance constrains the utility of speed. For example, under the tightest endurance at $N = 10$, the objective remains at $-45{,}243$ across all drone speeds. Since the flight radius is too restrictive to permit any drone sorties, the truck is forced to visit all collected customers alone. This bottleneck persists at larger scales: at $N = 100$, upgrading from a truck-only fleet to the fastest drone ($\phi = 3$) improves the objective by only $33{,}617$ under restricted endurance of $f = 0.25$, compared to a $61{,}863$ improvement under looser endurance of $f = 0.50$. Range and speed are therefore complementary; a faster drone only generates value when it has the battery capacity to reach its targets. 

\subsection{Parameter sensitivity}
\label{sec:ttd-oat}

A one-at-a-time sensitivity analysis explores variations from the \texttt{ttd300} baseline across four axes: the renting ratio $R \in \{1, 12.5, 25, 50, 100, 200\}$, capacity $W \in \{0.25, 0.5, 1, 2, 4\}\times$, minimum truck speed $v_{\min} \in \{0.05, 0.1, 0.25, 0.5, 0.9\}$, and drone speed factor $\phi \in \{0.5, 1, 2, 3, 4\}$. These are evaluated across all seven problem sizes and all four endurance fractions. Tables~S6 and~S7 in the Supplementary Material detail the results for different instance sizes.

The renting ratio dominates physical parameters, and this dominance strengthens as $N$ increases. At $N = 100$ and $f = 1.00$, varying $R$ yields objectives ranging from $+81{,}408$ at $R = 1$ down to $-382{,}171$ at $R = 200$. This depicts the impact of $R$. Simply lowering the renting ratio to $R = 25$ makes the operation profitable for $N \geq 40$, assuming $f \geq 0.5$, or from $N = 75$ at the tightest radius, without requiring any hardware upgrades to the fleet. Conversely, at $R = 200$, the algorithm abandons items at a large proportion of the visited customers, prioritising omissions along the truck route, because the cost of idling at rendezvous nodes dwarfs the potential profit of the items.

Among the physical parameters, capacity exerts the strongest influence, improving the objective from $-82{,}931$ to $-17{,}648$ at $N = 100$ and $f = 1.00$. This is followed by drone speed from $-57{,}656$ to $-23{,}416$ and minimum truck speed from $-41{,}311$ to $-18{,}705$. As anticipated, tightening the endurance fraction to $f = 0.25$ exclusively curtails the impact of drone speed: it compresses the performance range of $\phi$ at $N = 100$ to roughly $27{,}000$ and flattens it entirely at $N = 10$, while leaving the impacts of $R$, capacity, and $v_{\min}$ unchanged. The economic parameters, therefore, dictate the fundamental viability of the solution, while physical parameters merely provide marginal adjustments. Altering the renting ratio fundamentally changes the nature of the problem, whereas increasing capacity or speed offers incremental gains. Crucially, for an endurance-limited fleet, the value of a faster drone remains subordinate to its battery capacity.

\section{Conclusions}
\label{sec:conclusions}

This paper introduced the Travelling Thief Problem with Drone (TTP-D), a novel routing extension that couples load-dependent travel times and knapsack packing with time-synchronised, single-package drone sorties. To address the problem's inherent complexity, we developed an exact MILP formulation alongside a suite of scalable solution methods: SA and VNS metaheuristics, an attention-based DRL construction policy, and LISA, a hybrid algorithm. We evaluated these approaches on instances derived from the \texttt{a280} benchmark and a newly introduced, endurance-controlled dataset. 

Our computational study demonstrated that while the exact solver certifies optimality only for small instances up to $N = 10$, SA provides a robust reference for solution quality at larger scales. The learned policies construct solutions rapidly but struggle to match the quality of metaheuristic solutions on larger instances. However, the LISA bridges this gap by using the learned policy to warm-start a truncated search; it recovers near-metaheuristic performance at a fraction of the computational cost, although the largest instances still necessitate the full annealing budget.

From an operational perspective, our sensitivity analysis revealed that the time-based rental ratio overwhelmingly dictates overall mission profitability, overshadowing physical parameters such as vehicle capacity and speed. Drone endurance acts as a strict bottleneck; hence, logistics operators should prioritise extending operational range over increasing flight speed. Furthermore, allowing the collection of multiple items at each location can effectively transform large-scale, loss-making operations into profitable missions. These modelling insights highlight several natural avenues for future research, including incorporating multiple drones, multi-customer drone sorties, explicit handover times, and stochastic environments. Algorithmically, improving robust policy transfer across varying instance sizes and extending the optimality certification beyond $N = 20$ using exact methods remain vital open challenges.

\bibliographystyle{plainnat}
\bibliography{main}


\newpage
\begin{center}
    \LARGE \bfseries Supplementary Material for\\ \textit{Drive, Pack, Fly: The Travelling Thief Problem with Drone}
\end{center}
\vspace{1.5em}
This supplementary material comprises two main parts: 
Sections~S1--S4 contain constant tables, the derivation of the piecewise-linear error bound, an ablation study, and a seed-dispersion table; Section~S5 details the sensitivity analysis results, while Section~S6 provides the endurance reference for the \texttt{ttd300} benchmark.

\section*{S1. Constants of the model and the learning pipeline}

Table~\ref{tab:bigm} gives the disjunctive constants of the mixed-integer formulation, and Table~\ref{tab:hparams} lists the policy and training hyperparameters. Both support the content discussed and applied in the main manuscript. 

\begin{table}[htbp]
\centering
\caption{Value for big-$M$s. These constraint groups comprise the travel-time and synchronisation constraints presented in the manuscript.}
\label{tab:bigm}
\small
\begin{tabularx}{\textwidth}{@{} l c l X @{}}
\toprule
\textbf{Constraints} & \textbf{big-$M$} & \textbf{Tight Bound} & \textbf{Comment} \\
\midrule
Drone payload, weight propagation
  & $M_{\mathrm{W}}$
  & $w_{\mathrm{tot}} = \sum_{i \in \mathcal{N}} w_i$
  & Total weight of all available customer items. \\
\addlinespace
Truck timing
  & $M_{\mathrm{T}}$
  & $d_{\max} (N+1) / v_{\min}$
  & Worst-case traversal time assuming $N+1$ arcs at distance $d_{\max}$ and minimum velocity $v_{\min}$.\\
\addlinespace
Drone timing
  & $M_{\mathrm{D}}$
  & $d_{\mathrm{NN}} / v_{\min}$
  & Makespan of a nearest-neighbour warm-start tour evaluated at minimum velocity $v_{\min}$. \\
\bottomrule
\end{tabularx}
\end{table}

\begin{table}[htbp]
\centering
\caption{Policy and training hyperparameters. Here $d_{\mathrm{model}} = 128$ is the model dimension and
  $N$ is the number of customers.}
\label{tab:hparams}
\small
\renewcommand{\arraystretch}{1.15}
\begin{tabular}{@{}lll@{}}
\toprule
Group & Parameter & Value \\
\midrule
Encoder    & layers / model dim / heads / FF dim & $3$ / $128$ / $8$ / $512$ \\
           & activation / normalisation / dropout & GELU / pre-norm LayerNorm / $0$ \\
           & node features & coordinates, $p_i/p_{\max}$, $w_i/W$ \\
\midrule
Decoder    & pointer clipping constant $C$ & $10$ \\
           & state scalars & $7$ \\
           & binary heads & MLP$(2d_{\mathrm{model}} \to d_{\mathrm{model}} \to 2)$, GELU \\
\midrule
POMO       & rollouts per instance $P$ & $32$ \\
           & augmentations & $8$ (dihedral) \\
           & first-move strata & all feasible $(j_0, k_0)$ pairs \\
\midrule
PPO        & clip $\epsilon_{\mathrm{clip}}$ & $0.2$ \\
           & update epochs / minibatch & $3$ / $1{,}024$ steps \\
           & target KL (early stop) & $0.02$ (stop at $1.5\times$) \\
           & entropy coefficient & $0.02 \to 0.002$, cosine \\
           & advantage & group-centred, $\kappa_\mathcal{I}$-scaled, standardised \\
\midrule
Optimiser  & AdamW weight decay & $10^{-4}$ \\
           & peak / final LR / warmup & $10^{-4}$ / $2\times10^{-5}$ / $5$ epochs \\
           & gradient clip & $1.0$ \\
\midrule
Episode    & horizon cap & $2(N+2)$ epochs \\
           & seed & $2026$ \\
\midrule
Inference  & decoder & beam search, $8$ dihedral views \\
           & beam width & $256$, stratified by first launch \\
           & ablation budgets (Section~S3) & beam $512$; $128$ samples per (start, view) \\
\bottomrule
\end{tabular}
\end{table}

\paragraph{Implementation} The model and training pipeline are implemented in PyTorch~2.4.0, while the environment is built using NumPy. High accelerator utilisation during training is maintained via lockstep batched decoding. Training follows the shared configuration detailed in Table~\ref{tab:hparams}, with the random seed for instance sampling and weight initialisation fixed at $2026$. For each problem size, the checkpoint that achieves the lowest held-out gap during training is selected as the final solver.

\section*{S2. Error of the piecewise-linear travel time}

\begingroup
\ifmarkchanges\fi

Section 4.3 of the main paper replaces the reciprocal truck speed with an SOS2 chord interpolant and states, in constraint (18), a bound on the objective error introduced by this substitution. This section derives that bound.

\paragraph{Setting} 
Let the truck speed be a function of the load $u$ carried out of a node:
\begin{equation}
v(u) \;=\; v_{\max} - u\,\frac{\Delta v}{W}, \qquad
\psi(u) \;=\; \frac{1}{v(u)}, \qquad \Delta v = v_{\max} - v_{\min},
\label{eq:s-speed}
\end{equation}
so that in the exact problem, the truck spends $d_{ij}\psi(W_i)$ time units on arc $(i,j)$. The breakpoints are uniformly distributed across the realisable load range $[0, w_{\mathrm{tot}}]$. That is, $w^{\mathrm{bp}}_b = bh$ with spacing $h = w_{\mathrm{tot}}/K$ for $b \in \{0, \ldots, K\}$, and $v^{\mathrm{bp}}_b = v(w^{\mathrm{bp}}_b)$.

In the instances on which the model is solved, the total selectable weight does not exceed the capacity, $w_{\mathrm{tot}} \leq W$. Thus, the interpolation range lies strictly within $[0, W]$, and speeds remain positive throughout: $v(u) \geq v_{\min} > 0$, with equality occurring only at full capacity. Let $\hat\psi$ denote the piecewise-linear function that coincides with $\psi$ at the breakpoints and is linear on each interval $[w^{\mathrm{bp}}_b, w^{\mathrm{bp}}_{b+1}]$, and define
\begin{equation}
\epsilon_\psi \;=\; \max_{u \in [0, w_{\mathrm{tot}}]} \big( \hat\psi(u) - \psi(u) \big)
\label{eq:s-eps-psi}
\end{equation}
as the maximum error the interpolant introduces at any single node.

\begin{proposition}
Let $G^*_{\mathrm{MILP}}$ be the optimal value of the model in Section~4 of the main paper and $G^*_{\mathrm{exact}}$ the optimal value of the same model with $\hat\psi_i$ replaced by the exact reciprocal speed $\psi(W_i)$. Then

\begin{equation}
0 \;\leq\; G^*_{\mathrm{exact}} - G^*_{\mathrm{MILP}} \;\leq\; \epsilon_G
\;=\; R \cdot d_{\max} \cdot (N+1) \cdot
\left(\frac{w_{\mathrm{tot}}}{W}\right)^{\!2}
\frac{(\Delta v)^2}{4 K^2 v_{\min}^3}.
\label{eq:s-obj-error}
\end{equation}
\end{proposition}

The five steps below prove the proposition. Steps~1 to~3 bound the error the interpolant commits at a single node, Step~4 propagates it along a route, and Step~5 converts it into an objective error.

\paragraph{Step 1: The SOS2 system evaluates $\hat\psi$} Constraints~(14) and~(15)
of the main paper do not merely relax the reciprocal speed, they evaluate the interpolant exactly. The SOS2 condition permits at most two consecutive weights $\mu_{i,b}$ to be nonzero. If these are $\mu_{i,b} = 1 - \theta$ and $\mu_{i,b+1} = \theta$ with $\theta \in [0,1]$, then~(14) reads $W_i = (1-\theta)\,w^{\mathrm{bp}}_b + \theta\,w^{\mathrm{bp}}_{b+1}$ and $\hat\psi_i = (1-\theta)\,\psi(w^{\mathrm{bp}}_b) + \theta\,\psi(w^{\mathrm{bp}}_{b+1})$, which is the value at $W_i$ of the chord over $[w^{\mathrm{bp}}_b, w^{\mathrm{bp}}_{b+1}]$. Hence $\hat\psi_i = \hat\psi(W_i)$ for every feasible point, and the approximation error of the model is exactly the interpolation error of $\hat\psi$.

\paragraph{Step 2: The error is one-sided} Differentiating~\eqref{eq:s-speed}
twice with $a = \Delta v / W$ gives
\begin{equation}
\psi'(u) \;=\; \frac{a}{v(u)^2}, \qquad
\psi''(u) \;=\; \frac{2a^2}{v(u)^3} \;>\; 0,
\label{eq:s-derivs}
\end{equation}
so $\psi$ is strictly convex and increasing on $[0, W]$. A chord of a convex function lies above it, so $\hat\psi(u) \geq \psi(u)$ on the whole range, with equality at the break-points. The model therefore never underestimates a truck travel time: $\hat\psi_i \geq 1/v_i$, and $\epsilon_\psi \geq 0$ in~\eqref{eq:s-eps-psi}.

\paragraph{Step 3: The magnitude of the per-node error} On a given interval
$[w^{\mathrm{bp}}_b, w^{\mathrm{bp}}_{b+1}]$ of length $h$, the error of linear interpolation at the two endpoints is
\begin{equation}
\hat\psi(u) - \psi(u) \;=\; -\tfrac{1}{2}\,\psi''(\xi)\,
\big(u - w^{\mathrm{bp}}_b\big)\big(u - w^{\mathrm{bp}}_{b+1}\big)
\end{equation}
for some $\xi$ in the interval. The quadratic factor is at most $h^2/4$ in absolute value, attained at the midpoint, so $\hat\psi(u) - \psi(u) \leq \tfrac{1}{8} h^2 \max_u \psi''(u)$. By~\eqref{eq:s-derivs} the second derivative is increasing in $u$ and is bounded on the realisable range by its value at full capacity, $\psi''(u) \leq 2 (\Delta v / W)^2 / v_{\min}^3$. With $h = w_{\mathrm{tot}}/K$,
\begin{equation}
\epsilon_\psi \;\leq\;
\frac{1}{8} \left(\frac{w_{\mathrm{tot}}}{K}\right)^{\!2}
\frac{2 (\Delta v / W)^2}{v_{\min}^3}
\;=\;
\left(\frac{w_{\mathrm{tot}}}{W}\right)^{\!2}
\frac{(\Delta v)^2}{4 K^2 v_{\min}^3},
\label{eq:s-pl-error}
\end{equation}
which decreases as $O(K^{-2})$. Two of the three steps here are conservative: the midpoint value is attained on at most one interval, and $\psi''$ reaches its bound only when the truck is loaded to capacity, which no realisable load does once $w_{\mathrm{tot}} < W$. Replacing $v_{\min}$ by $v(w_{\mathrm{tot}})$, the slowest speed an instance can actually reach, gives a tighter constant; constraint~(18) keeps $v_{\min}$ so that the bound depends only on the model parameters.

\paragraph{Step 4: Propagation into the makespan} Fix any routing, assignment and packing $(x, y, z)$ that satisfies constraints~(2)--(13) of the main paper. Because none of those constraints involves $\psi$, the departing loads $W_i$ remain identical in both models. Consequently, by Step 1, the induced values $\hat\psi(W_i)$ are also identical. 
Let $\pi_0, \pi_1, \ldots, \pi_m$ (where $\pi_0$ is the start depot and $\pi_m$ is the end depot) represent the truck path, comprising $m \leq N+1$ arcs. Let $\tau$ and $\hat\tau$ denote the earliest arrival times consistent with the timing constraints (16)–(17) of the main paper under $\psi$ and $\hat\psi$, respectively. Both satisfy the same recursion along the path:
\begin{equation}
\tau_{\pi_r} \;=\; \max\left\{
\tau_{\pi_{r-1}} + d_{\pi_{r-1}\pi_r}\,\psi(W_{\pi_{r-1}}),\;
\max_{(\ell, k) \,:\, \text{sortie landing at } \pi_r}
\tau_{\ell} + \frac{d_{\ell k} + d_{k \pi_r}}{v_{\mathrm{D}}}
\right\}.
\label{eq:s-recursion}
\end{equation}

The drone terms are identical in the two systems because the drone flies at the constant speed $v_{\mathrm{D}}$. 
Let $\delta_r = \hat\tau_{\pi_r} - \tau_{\pi_r}$. Since $\delta_0 = 0$, every launch node $\ell$ precedes its rendezvous node on the truck path, and a maximum of finitely many terms move by at most the largest move of any term, we have:
\begin{equation}
\delta_r \;\leq\; \max\Big\{ \delta_{r-1} + d_{\pi_{r-1}\pi_r}\,\epsilon_\psi,\;
\max_{\ell \prec \pi_r} \delta_\ell \Big\}.
\end{equation}

Induction on $r$ gives $\delta_r \leq \epsilon_\psi \sum_{s \leq r} d_{\pi_{s-1}\pi_s}$, and with Step~2 for the lower end,
\begin{equation}
0 \;\leq\; \hat\tau_{\depotend} - \tau_{\depotend}
\;\leq\; d_{\max} (N+1)\, \epsilon_\psi .
\label{eq:s-makespan-error}
\end{equation}

This follows the exact same accounting logic used to establish the disjunctive constant $M_{\mathrm{T}}$ in Table~\ref{tab:bigm}: at most $N+1$ truck arcs, each of length at most $d_{\max}$.

\paragraph{Step 5: From makespan to objective} Those two constraints bound $\tau$ from below only, and the objective~(1) of the main paper is decreasing in $\tau_{\depotend}$, so at an optimal solution $\tau_{\depotend}$ takes the earliest feasible value used in Step~4. No constraint in the model restricts $\tau$ from above, and the sortie endurance limit of the ttd300 benchmark caps a round-trip distance rather than a time. Thus, the timing variables never dictate feasibility: both models share the exact same feasible set in terms of $(x, y, z)$, differing only in the objective values they assign to these solutions. Since the collected profit $\sum_i p_i z_i$ is unaffected, \eqref{eq:s-makespan-error} gives, for every feasible $(x, y, z)$,
\begin{equation}
0 \;\leq\; \Big( G_{\mathrm{exact}}(x,y,z) - G_{\mathrm{MILP}}(x,y,z) \Big)
\;=\; R \big( \hat\tau_{\depotend} - \tau_{\depotend} \big)
\;\leq\; R\, d_{\max} (N+1)\, \epsilon_\psi .
\end{equation}

Maximising the left-hand relation over the common feasible set yields $G^*_{\mathrm{MILP}} \leq G^*_{\mathrm{exact}}$, and evaluating the right-hand one at an exact optimiser yields $G^*_{\mathrm{MILP}} \geq G^*_{\mathrm{exact}} - R\, d_{\max} (N+1)\,\epsilon_\psi$. Substituting~\eqref{eq:s-pl-error} for $\epsilon_\psi$ gives~\eqref{eq:s-obj-error}, which is constraint~(18) of the main paper. \hfill$\square$

\paragraph{Reading of the bound} The approximation is conservative in one
direction: the MILP optimum is a valid lower bound on the exact optimum, and a solution certified optimal for the model is within $\epsilon_G$ of the exact optimum. The bound is worst-case in every factor, since it charges the maximum interpolation error to every arc, the maximum arc length to every truck leg, and the full-capacity speed to every load. The factor $v_{\min}^{-3}$ dominates it and is attained only at capacity, so on instances whose collected weight stays well below $W$, the realised error is substantially smaller than the bound. Nevertheless, this residual error explains the discrepancy between the two objectives in the computational experiments: the model is optimised on the interpolant at $K = 10$ break-points while every reported objective is evaluated on the exact load--speed law, which is why the MILP values at $N = 15$ in Section~6.1.3 of the main paper sit below the annealing optimum even though the model is solved to optimality.

\endgroup

\section*{S3. Decode-strategy ablation}

\begin{figure}[htbp]
\centering
\includegraphics[width=\textwidth]{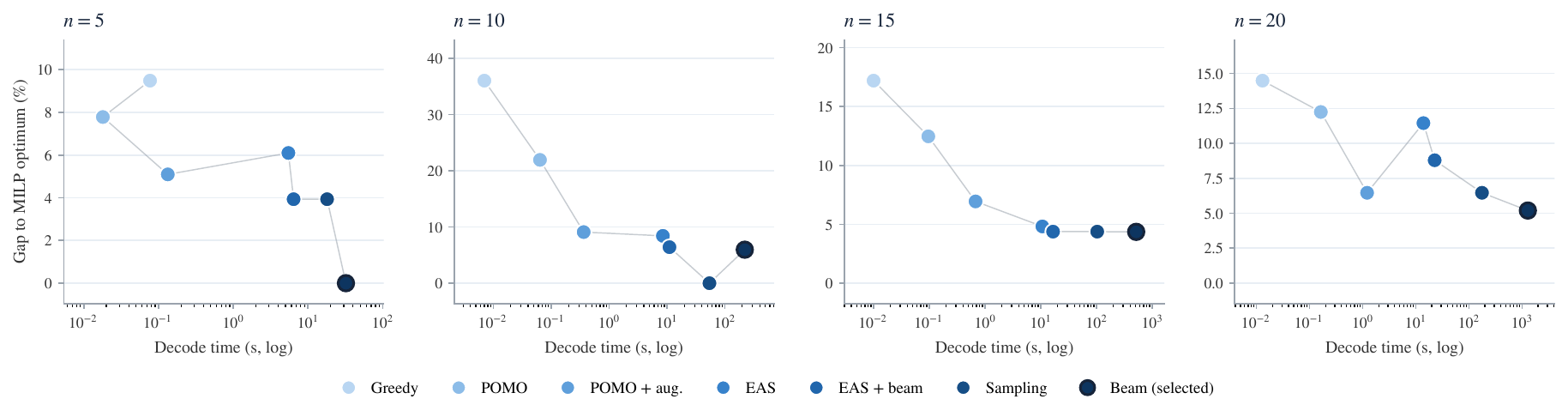}
\caption{Decode-strategy ablation on the a280 MILP-anchored instance sizes. Each panel plots the gap to the certified MILP optimum against decode runtime (log scale) for seven decoding strategies under one trained policy, one run per strategy. The strategies form a ladder of increasing decode compute, shaded light to dark in that order, with a connector following the ladder within each panel and a ring on the selected decoder.}
\label{fig:decode_ablation}
\end{figure}

This section details the selection of the decoder used by the learned policy in the main paper. We ultimately select beam search from a progression of seven strategies with increasing computational requirements. These span a single greedy rollout; the POMO multi-start with and without dihedral augmentation; temperature-$1$ sampling applied to the augmented multi-start; a stratified beam search; and efficient active search \citep{hottung2022eas}, evaluated both on its own and combined with the beam. We compare these approaches on the same foundation that anchors the rest of the study: the \texttt{a280}-derived instances at $N \in \{5, 10, 15, 20\}$. Because the certified MILP optimum provides an exact reference for these instances, each strategy's optimality gap is measured against a proven target rather than a heuristic baseline. Figure~\ref{fig:decode_ablation} plots the resulting optimality gap against the decoding runtime for each instance size.

The pattern is consistent across the four instance sizes. The least expensive strategies leave the largest gaps: greedy decoding trails the optimum by $9.5\%$ to $36.0\%$ and the plain POMO multi-start by $7.8\%$ to $21.9\%$ with augmentation, while sampling closes much of that distance as the budget grows. Beam search attains the lowest gap at three of the four sizes and is the only strategy that remains near the frontier at every size, whereas each of the cheaper alternatives gives up several percentage points at one size or another. It is also the most expensive decoder, an order of magnitude above sampling and more than that above the multi-start variants. We accept this cost because the per-instance decode remains far below the metaheuristic budgets at the sizes where the policy is used, and because the resulting quality is what the LISA hybrid distils into its warm start.

\section*{S4. Seed dispersion on the a280 benchmark}

\begin{table}[h]
\centering
\caption{Seed-to-seed dispersion of the objective (\%) on the \texttt{a280}
  benchmark.}
\label{tab:dispersion}
\renewcommand{\arraystretch}{1.2}
\setlength{\tabcolsep}{6pt}
\small
\begin{tabular}{@{}lccccccc@{}}
\toprule
Method & $N=5$ & $N=10$ & $N=15$ & $N=20$ & $N=30$ & $N=40$ & $N=50$ \\
\midrule
SA  & 0.00 & 0.33 & 0.76 & 0.94 & 1.98 & 1.94 & 2.34 \\
VNS & 0.01 & 0.00 & 0.42 & 3.02 & 5.04 & 5.49 & 4.96 \\
\bottomrule
\end{tabular}
\end{table}

Mean gaps alone can hide run-to-run variability.
Table~\ref{tab:dispersion} reports, for the two per-instance searches, the standard deviation of the objective across the ten seeds on each instance, expressed as a percentage of that instance's mean absolute objective and averaged over the five instances per size. Simulated annealing is concentrated as well as strong: its dispersion stays below $1\%$ of the objective through $N = 20$ and below $2.4\%$ through $N = 50$. VNS matches this concentration only through $N = 15$ and is roughly three times more dispersed from $N = 20$ through $N = 50$. From $N = 40$ onward, the difference between the SA and VNS mean gaps in the main paper exceeds the sum of the two standard deviations, so the ranking between the two searches at those sizes is not a seed artefact. Through $N = 30$, it does not, and the two searches should be read as close rather than as separate at those sizes, which is consistent with their gaps being within a third of a percentage point of one another through $N = 20$. As an exact method, the MILP is deterministic and naturally exempt from this seed-based variance.

\section*{S5. Details of the sensitivity analysis}

All sensitivity table entries are generated using the simulated-annealing solver (Section~5.1.1) under \texttt{ttd300} per-size budgets across all seven instance sizes $N \in \{10, 20, 30, 40, 50, 75, 100\}$, layout L1, and four endurance fractions $f = E_{\mathrm{D}}/d_{\max} \in \{0.25, 0.5, 0.75, 1.0\}$. Every plan is validated by the exact endurance-aware evaluator before scoring. Notation, benchmark settings, and the maximised objective $G = \sum_{i} p_i z_i - R\,\tau_{\depotend}$ (where less negative is better) follow the main paper. 
To limit computational cost, experiments are restricted to layout L1 and a single seed execution, except Table~\ref{tab:ttd-multi-item}, which spans five seeds to better navigate the larger multi-item search space.

Table~\ref{tab:ttd-multi-item} reports the multi-item comparison summarised in Section~7.1 of the main paper, Table~\ref{tab:ttd-drone-speed} the drone-speed sweep summarised in Section~7.2, and Tables~\ref{tab:ttd-oat-a} and~\ref{tab:ttd-oat-b} the one-at-a-time parameter sweep summarised in Section~7.3. 
Minor deviations from general trends in larger instances ($N \ge 50$) reflect standard metaheuristic behaviour, as a fixed runtime budget spans a diminishing share of the expanding solution space.

\begin{table}[htbp]\centering
\setlength{\tabcolsep}{4.5pt}
\renewcommand{\arraystretch}{1.1}
\caption{Performance comparison of multi-item versus single-item collection strategies on the \texttt{ttd300} benchmark. Each row reports the best result across independent runs with the same per-size budget on the first layout (L1), across varying instance sizes $N$ and drone endurance fractions $f = E_{\mathrm{D}}/d_{\max}$. The objective values for the multi-item and single-item models are denoted by $G_{\mathrm{MI}}$ and $G_{\mathrm{SI}}$, respectively, alongside their exact difference $\Delta G = G_{\mathrm{MI}} - G_{\mathrm{SI}}$. Operational metrics include the total number of collected cities (``Cities''), the average (``Avg'') and maximum (``Max'') number of items retrieved per collected city, the capacity utilisation fraction (``Util''), and the split of collected cities between the truck and the drone (``Tr''/``Dr''). The better objective value for each configuration is marked in bold.}
\label{tab:ttd-multi-item}\vspace{4pt}
\scriptsize
\resizebox{\tabfit}{!}{%
\begin{tabular}{@{}r r r r r r r r r r r@{}}
\toprule
$N$ & $f$ & $G_{\mathrm{MI}}$ & $G_{\mathrm{SI}}$ & $\Delta G$ & Cities & Avg & Max & Util & Tr & Dr \\
\midrule
\multirow{4}{*}{10} & 0.25 & \textbf{$-$43{,}618} & $-$45{,}243 & 1{,}626 & 4.0 & 2.50 & 4.0 & 44\% & 4.0 & 0.0 \\
 & 0.50 & \textbf{$-$37{,}994} & $-$41{,}253 & 3{,}259 & 4.0 & 3.25 & 5.0 & 57\% & 2.0 & 2.0 \\
 & 0.75 & \textbf{$-$30{,}665} & $-$34{,}623 & 3{,}958 & 6.0 & 2.83 & 5.0 & 75\% & 3.0 & 3.0 \\
 & 1.00 & \textbf{$-$25{,}665} & $-$29{,}670 & 4{,}004 & 7.0 & 2.57 & 5.0 & 80\% & 3.0 & 4.0 \\
\midrule
\multirow{4}{*}{20} & 0.25 & \textbf{$-$46{,}906} & $-$54{,}905 & 7{,}998 & 13.0 & 2.85 & 5.0 & 82\% & 8.0 & 5.0 \\
 & 0.50 & \textbf{$-$38{,}310} & $-$48{,}705 & 10{,}395 & 13.0 & 2.92 & 5.0 & 84\% & 8.0 & 5.0 \\
 & 0.75 & \textbf{$-$21{,}186} & $-$28{,}912 & 7{,}727 & 18.0 & 2.44 & 5.0 & 97\% & 10.0 & 8.0 \\
 & 1.00 & \textbf{$-$17{,}677} & $-$27{,}925 & 10{,}248 & 16.0 & 2.69 & 5.0 & 95\% & 8.0 & 8.0 \\
\midrule
\multirow{4}{*}{30} & 0.25 & \textbf{$-$41{,}442} & $-$52{,}774 & 11{,}331 & 18.0 & 2.78 & 5.0 & 74\% & 11.0 & 7.0 \\
 & 0.50 & \textbf{$-$25{,}796} & $-$35{,}270 & 9{,}474 & 24.0 & 2.29 & 5.0 & 81\% & 14.0 & 10.0 \\
 & 0.75 & \textbf{$-$17{,}239} & $-$30{,}404 & 13{,}164 & 26.0 & 2.35 & 5.0 & 90\% & 15.0 & 11.0 \\
 & 1.00 & \textbf{$-$16{,}512} & $-$27{,}842 & 11{,}330 & 22.0 & 2.77 & 5.0 & 90\% & 13.0 & 9.0 \\
\midrule
\multirow{4}{*}{40} & 0.25 & \textbf{$-$31{,}319} & $-$43{,}502 & 12{,}182 & 27.0 & 2.41 & 5.0 & 72\% & 19.0 & 8.0 \\
 & 0.50 & \textbf{$-$17{,}425} & $-$32{,}411 & 14{,}986 & 29.0 & 2.48 & 5.0 & 80\% & 16.0 & 13.0 \\
 & 0.75 & \textbf{$-$15{,}693} & $-$32{,}453 & 16{,}760 & 32.0 & 2.53 & 5.0 & 89\% & 21.0 & 11.0 \\
 & 1.00 & \textbf{$-$13{,}652} & $-$31{,}374 & 17{,}721 & 30.0 & 2.60 & 5.0 & 86\% & 20.0 & 10.0 \\
\midrule
\multirow{4}{*}{50} & 0.25 & \textbf{$-$22{,}944} & $-$45{,}629 & 22{,}685 & 38.0 & 2.71 & 5.0 & 91\% & 24.0 & 14.0 \\
 & 0.50 & \textbf{$-$5{,}336} & $-$30{,}438 & 25{,}102 & 40.0 & 2.75 & 5.0 & 97\% & 24.0 & 16.0 \\
 & 0.75 & \textbf{$-$5{,}214} & $-$29{,}955 & 24{,}741 & 42.0 & 2.69 & 5.0 & 100\% & 24.0 & 18.0 \\
 & 1.00 & \textbf{$-$4{,}943} & $-$29{,}701 & 24{,}758 & 41.0 & 2.61 & 5.0 & 95\% & 24.0 & 17.0 \\
\midrule
\multirow{4}{*}{75} & 0.25 & \textbf{$-$12{,}795} & $-$39{,}164 & 26{,}370 & 59.0 & 2.42 & 5.0 & 84\% & 37.0 & 22.0 \\
 & 0.50 & \textbf{$-$6{,}234} & $-$32{,}639 & 26{,}405 & 61.0 & 2.43 & 5.0 & 87\% & 38.0 & 23.0 \\
 & 0.75 & \textbf{$-$3{,}790} & $-$32{,}696 & 28{,}905 & 61.0 & 2.49 & 5.0 & 90\% & 42.0 & 19.0 \\
 & 1.00 & \textbf{$-$3{,}981} & $-$31{,}554 & 27{,}573 & 62.0 & 2.47 & 5.0 & 90\% & 43.0 & 19.0 \\
\midrule
\multirow{4}{*}{100} & 0.25 & \textbf{7{,}279} & $-$44{,}719 & 51{,}998 & 77.0 & 2.64 & 5.0 & 90\% & 56.0 & 21.0 \\
 & 0.50 & \textbf{5{,}259} & $-$34{,}799 & 40{,}058 & 79.0 & 2.63 & 5.0 & 92\% & 55.0 & 24.0 \\
 & 0.75 & \textbf{6{,}864} & $-$40{,}610 & 47{,}474 & 84.0 & 2.44 & 5.0 & 90\% & 60.0 & 24.0 \\
 & 1.00 & \textbf{13{,}342} & $-$40{,}603 & 53{,}945 & 86.0 & 2.47 & 5.0 & 94\% & 60.0 & 26.0 \\
\bottomrule\end{tabular}}
\setlength{\tabcolsep}{6pt}
\end{table}

\begin{table}[htbp]\centering
\setlength{\tabcolsep}{4pt}
\renewcommand{\arraystretch}{1.1}
\caption{Drone speed sensitivity analysis on the \texttt{ttd300} benchmark. For each instance size $N$ and speed factor $\phi = v_{\mathrm{D}} / v_{\max}$ (where $\phi = 0$ establishes the truck-only baseline), each endurance-fraction column group reports the objective value $G$ on the first layout (L1), alongside the division of collected cities across the truck, drone, and rendezvous points. The best objective value per $(N, f)$ block is highlighted in bold.}
\label{tab:ttd-drone-speed}\vspace{4pt}
\scriptsize
\resizebox{\tabfit}{!}{%
\begin{tabular}{@{}r r r c r c r c r c@{}}
\toprule
 & & \multicolumn{2}{c}{$f = 0.25$} & \multicolumn{2}{c}{$f = 0.50$} & \multicolumn{2}{c}{$f = 0.75$} & \multicolumn{2}{c}{$f = 1.00$} \\
\cmidrule(lr){3-4} \cmidrule(lr){5-6} \cmidrule(lr){7-8} \cmidrule(lr){9-10}
$N$ & $\phi$ & $G$ & T/D/C & $G$ & T/D/C & $G$ & T/D/C & $G$ & T/D/C \\
\midrule
\multirow{5}{*}{10} & 0 & \textbf{$-$45{,}243} & 5.0/0.0/0.0 & $-$45{,}243 & 5.0/0.0/0.0 & $-$46{,}738 & 3.0/0.0/0.0 & $-$45{,}819 & 5.0/0.0/0.0 \\
 & 0.5 & \textbf{$-$45{,}243} & 5.0/0.0/0.0 & $-$43{,}646 & 4.0/1.0/1.0 & $-$41{,}515 & 4.0/1.0/2.0 & $-$40{,}142 & 5.0/1.0/1.0 \\
 & 1 & \textbf{$-$45{,}243} & 5.0/0.0/0.0 & $-$42{,}123 & 5.0/2.0/2.0 & $-$38{,}215 & 5.0/4.0/4.0 & $-$36{,}644 & 5.0/3.0/4.0 \\
 & 2 & \textbf{$-$45{,}243} & 5.0/0.0/0.0 & \textbf{$-$41{,}253} & 3.0/2.0/2.0 & $-$34{,}623 & 4.0/3.0/4.0 & $-$29{,}670 & 4.0/5.0/5.0 \\
 & 3 & \textbf{$-$45{,}243} & 5.0/0.0/0.0 & \textbf{$-$41{,}253} & 3.0/2.0/2.0 & \textbf{$-$33{,}877} & 3.0/3.0/4.0 & \textbf{$-$28{,}486} & 4.0/3.0/5.0 \\
\midrule
\multirow{5}{*}{20} & 0 & $-$67{,}029 & 8.0/0.0/0.0 & $-$67{,}029 & 8.0/0.0/0.0 & $-$67{,}029 & 8.0/0.0/0.0 & $-$67{,}029 & 8.0/0.0/0.0 \\
 & 0.5 & $-$60{,}872 & 12.0/1.0/1.0 & $-$57{,}580 & 11.0/3.0/3.0 & $-$50{,}336 & 13.0/2.0/2.0 & $-$48{,}908 & 13.0/3.0/3.0 \\
 & 1 & $-$58{,}263 & 11.0/5.0/6.0 & $-$52{,}122 & 10.0/5.0/6.0 & $-$47{,}050 & 15.0/5.0/5.0 & $-$41{,}646 & 13.0/7.0/7.0 \\
 & 2 & $-$54{,}905 & 10.0/5.0/6.0 & $-$48{,}705 & 9.0/7.0/8.0 & $-$29{,}439 & 11.0/9.0/9.0 & $-$27{,}953 & 14.0/6.0/6.0 \\
 & 3 & \textbf{$-$54{,}738} & 10.0/5.0/6.0 & \textbf{$-$47{,}410} & 9.0/8.0/8.0 & \textbf{$-$24{,}857} & 11.0/9.0/9.0 & \textbf{$-$22{,}423} & 10.0/10.0/10.0 \\
\midrule
\multirow{5}{*}{30} & 0 & $-$65{,}791 & 21.0/0.0/0.0 & $-$69{,}373 & 16.0/0.0/0.0 & $-$69{,}373 & 16.0/0.0/0.0 & $-$69{,}373 & 16.0/0.0/0.0 \\
 & 0.5 & $-$59{,}666 & 20.0/2.0/3.0 & $-$55{,}512 & 18.0/4.0/4.0 & $-$54{,}492 & 19.0/3.0/4.0 & $-$51{,}687 & 23.0/2.0/3.0 \\
 & 1 & $-$57{,}224 & 18.0/6.0/8.0 & $-$47{,}023 & 20.0/7.0/8.0 & $-$44{,}406 & 20.0/7.0/7.0 & $-$40{,}681 & 22.0/6.0/7.0 \\
 & 2 & $-$52{,}774 & 18.0/7.0/9.0 & $-$36{,}715 & 16.0/13.0/13.0 & $-$30{,}404 & 18.0/11.0/11.0 & $-$29{,}843 & 16.0/14.0/14.0 \\
 & 3 & \textbf{$-$51{,}843} & 18.0/7.0/9.0 & \textbf{$-$33{,}952} & 15.0/14.0/14.0 & \textbf{$-$24{,}689} & 15.0/14.0/15.0 & \textbf{$-$23{,}336} & 16.0/13.0/14.0 \\
\midrule
\multirow{5}{*}{40} & 0 & $-$61{,}254 & 31.0/0.0/0.0 & $-$63{,}498 & 31.0/0.0/0.0 & $-$63{,}498 & 31.0/0.0/0.0 & $-$63{,}498 & 31.0/0.0/0.0 \\
 & 0.5 & $-$55{,}520 & 34.0/3.0/4.0 & $-$52{,}150 & 32.0/4.0/5.0 & $-$50{,}573 & 33.0/4.0/4.0 & $-$51{,}382 & 33.0/4.0/4.0 \\
 & 1 & $-$49{,}771 & 26.0/11.0/12.0 & $-$42{,}051 & 27.0/9.0/11.0 & $-$41{,}720 & 28.0/8.0/10.0 & $-$40{,}532 & 27.0/10.0/10.0 \\
 & 2 & $-$43{,}626 & 24.0/12.0/14.0 & $-$32{,}699 & 23.0/16.0/16.0 & $-$32{,}804 & 23.0/17.0/17.0 & $-$32{,}515 & 24.0/16.0/16.0 \\
 & 3 & \textbf{$-$42{,}679} & 22.0/14.0/16.0 & \textbf{$-$30{,}772} & 21.0/19.0/19.0 & \textbf{$-$28{,}264} & 22.0/16.0/17.0 & \textbf{$-$27{,}742} & 22.0/17.0/17.0 \\
\midrule
\multirow{5}{*}{50} & 0 & $-$76{,}864 & 40.0/0.0/0.0 & $-$77{,}069 & 40.0/0.0/0.0 & $-$78{,}498 & 39.0/0.0/0.0 & $-$76{,}519 & 40.0/0.0/0.0 \\
 & 0.5 & $-$56{,}178 & 40.0/6.0/6.0 & $-$54{,}751 & 41.0/4.0/5.0 & $-$53{,}983 & 39.0/5.0/5.0 & $-$53{,}072 & 40.0/6.0/6.0 \\
 & 1 & $-$50{,}965 & 36.0/13.0/13.0 & $-$40{,}610 & 38.0/10.0/10.0 & $-$41{,}499 & 36.0/12.0/12.0 & $-$41{,}915 & 39.0/10.0/10.0 \\
 & 2 & $-$45{,}767 & 30.0/18.0/18.0 & $-$27{,}173 & 29.0/21.0/21.0 & $-$27{,}166 & 32.0/18.0/18.0 & $-$27{,}858 & 31.0/18.0/18.0 \\
 & 3 & \textbf{$-$42{,}851} & 30.0/19.0/19.0 & \textbf{$-$27{,}115} & 29.0/20.0/20.0 & \textbf{$-$24{,}488} & 27.0/23.0/23.0 & \textbf{$-$21{,}902} & 31.0/19.0/19.0 \\
\midrule
\multirow{5}{*}{75} & 0 & $-$66{,}536 & 65.0/0.0/0.0 & $-$80{,}272 & 68.0/0.0/0.0 & $-$80{,}434 & 68.0/0.0/0.0 & $-$80{,}272 & 68.0/0.0/0.0 \\
 & 0.5 & $-$58{,}583 & 58.0/10.0/10.0 & $-$56{,}289 & 60.0/9.0/9.0 & $-$54{,}266 & 65.0/8.0/8.0 & $-$55{,}126 & 61.0/7.0/7.0 \\
 & 1 & $-$50{,}757 & 52.0/19.0/19.0 & $-$46{,}723 & 58.0/14.0/15.0 & $-$45{,}283 & 58.0/15.0/15.0 & $-$46{,}100 & 56.0/15.0/15.0 \\
 & 2 & $-$42{,}880 & 42.0/30.0/30.0 & $-$33{,}963 & 47.0/26.0/26.0 & $-$34{,}561 & 52.0/23.0/23.0 & $-$31{,}554 & 53.0/22.0/22.0 \\
 & 3 & \textbf{$-$41{,}333} & 41.0/29.0/31.0 & \textbf{$-$26{,}822} & 43.0/30.0/30.0 & \textbf{$-$27{,}906} & 46.0/26.0/28.0 & \textbf{$-$27{,}111} & 46.0/26.0/27.0 \\
\midrule
\multirow{5}{*}{100} & 0 & $-$74{,}047 & 96.0/0.0/0.0 & $-$89{,}453 & 90.0/0.0/0.0 & $-$86{,}893 & 89.0/0.0/0.0 & $-$88{,}827 & 90.0/0.0/0.0 \\
 & 0.5 & $-$65{,}018 & 82.0/10.0/10.0 & $-$62{,}420 & 85.0/9.0/9.0 & $-$61{,}686 & 86.0/9.0/10.0 & $-$57{,}664 & 83.0/10.0/10.0 \\
 & 1 & $-$55{,}172 & 78.0/19.0/20.0 & $-$51{,}200 & 81.0/18.0/18.0 & $-$51{,}362 & 83.0/15.0/16.0 & $-$50{,}634 & 85.0/13.0/13.0 \\
 & 2 & $-$44{,}720 & 66.0/31.0/31.0 & $-$36{,}374 & 68.0/31.0/31.0 & \textbf{$-$32{,}453} & 72.0/25.0/25.0 & $-$38{,}667 & 67.0/31.0/31.0 \\
 & 3 & \textbf{$-$40{,}430} & 69.0/29.0/30.0 & \textbf{$-$27{,}590} & 66.0/33.0/34.0 & $-$35{,}352 & 65.0/35.0/35.0 & \textbf{$-$34{,}006} & 63.0/37.0/37.0 \\
\bottomrule\end{tabular}}
\setlength{\tabcolsep}{6pt}
\end{table}

\begin{table}[htbp]\centering
\setlength{\tabcolsep}{4pt}
\renewcommand{\arraystretch}{1.05}
\caption{One-at-a-time sensitivity analysis on the \texttt{ttd300} benchmark across all four endurance fractions, Part~1 ($R$ and $W$). Each parameter block evaluates variations from a shared baseline configuration ($R = 50$, $W = 1\times$ nominal capacity, $v_{\min} = 0.1$, $\phi = 2$). Within each tested parameter value, the four rows correspond to the endurance fractions $f = E_{\mathrm{D}}/d_{\max} \in \{0.25, 0.5, 0.75, 1.0\}$. Every entry details the objective value $G$ for that size, obtained from a single run on the first layout (L1). The highest $G$ in each (block, size, $f$) subset is in bold. Part~2 (Table~\ref{tab:ttd-oat-b}) reports on $v_{\min}$ and $\phi$.}
\label{tab:ttd-oat-a}\vspace{4pt}
\scriptsize
\resizebox{\tabfit}{!}{%
\begin{tabular}{@{}l l c r r r r r r r@{}}
\toprule
Param & Value & $f$ & $N=10$ & $N=20$ & $N=30$ & $N=40$ & $N=50$ & $N=75$ & $N=100$ \\
\midrule
\multirow{4}{*}{baseline} & \multirow{4}{*}{---}
 & 0.25 & $-$45{,}243 & $-$54{,}905 & $-$52{,}774 & $-$43{,}626 & $-$45{,}767 & $-$43{,}424 & $-$44{,}720 \\
 & & 0.50 & $-$41{,}253 & $-$48{,}705 & $-$36{,}715 & $-$32{,}699 & $-$27{,}221 & $-$33{,}963 & $-$38{,}322 \\
 & & 0.75 & $-$34{,}623 & $-$29{,}439 & $-$30{,}404 & $-$32{,}804 & $-$27{,}166 & $-$34{,}561 & $-$32{,}453 \\
 & & 1.00 & $-$29{,}670 & $-$27{,}953 & $-$29{,}843 & $-$32{,}515 & $-$27{,}858 & $-$31{,}554 & $-$38{,}929 \\
\midrule
\multirow{24}{*}{$R$}
 & \multirow{4}{*}{1}
 & 0.25 & \textbf{7{,}415} & \textbf{16{,}039} & \textbf{23{,}626} & \textbf{32{,}842} & \textbf{38{,}891} & \textbf{56{,}857} & \textbf{81{,}349} \\
 & & 0.50 & \textbf{7{,}507} & \textbf{16{,}178} & \textbf{24{,}009} & \textbf{33{,}063} & \textbf{39{,}240} & \textbf{57{,}062} & \textbf{81{,}543} \\
 & & 0.75 & \textbf{7{,}683} & \textbf{16{,}592} & \textbf{24{,}121} & \textbf{33{,}062} & \textbf{39{,}266} & \textbf{57{,}020} & \textbf{81{,}458} \\
 & & 1.00 & \textbf{7{,}802} & \textbf{16{,}615} & \textbf{24{,}125} & \textbf{33{,}078} & \textbf{39{,}201} & \textbf{56{,}993} & \textbf{81{,}408} \\
\cmidrule(l){2-10}
 & \multirow{4}{*}{12.5}
 & 0.25 & $-$5{,}899 & $-$1{,}032 & 5{,}434 & 14{,}819 & 19{,}158 & 33{,}632 & 51{,}734 \\
 & & 0.50 & $-$4{,}749 & 717 & 10{,}067 & 17{,}339 & 22{,}886 & 35{,}477 & 53{,}838 \\
 & & 0.75 & $-$2{,}550 & 5{,}914 & 11{,}033 & 17{,}568 & 23{,}401 & 35{,}569 & 53{,}300 \\
 & & 1.00 & $-$1{,}066 & 6{,}170 & 11{,}771 & 17{,}573 & 22{,}839 & 35{,}607 & 53{,}130 \\
\cmidrule(l){2-10}
 & \multirow{4}{*}{25}
 & 0.25 & $-$19{,}668 & $-$19{,}588 & $-$14{,}731 & $-$5{,}074 & $-$2{,}926 & 8{,}268 & 21{,}036 \\
 & & 0.50 & $-$17{,}586 & $-$16{,}089 & $-$5{,}437 & 120 & 5{,}911 & 13{,}079 & 22{,}567 \\
 & & 0.75 & $-$13{,}620 & $-$5{,}764 & $-$2{,}524 & 486 & 5{,}605 & 12{,}149 & 21{,}423 \\
 & & 1.00 & $-$10{,}705 & $-$5{,}215 & $-$2{,}118 & 1{,}304 & 5{,}342 & 11{,}930 & 21{,}429 \\
\cmidrule(l){2-10}
 & \multirow{4}{*}{50}
 & 0.25 & $-$45{,}243 & $-$54{,}905 & $-$52{,}774 & $-$43{,}626 & $-$44{,}978 & $-$43{,}424 & $-$44{,}720 \\
 & & 0.50 & $-$41{,}253 & $-$48{,}705 & $-$36{,}715 & $-$32{,}699 & $-$27{,}173 & $-$33{,}963 & $-$36{,}374 \\
 & & 0.75 & $-$34{,}623 & $-$29{,}439 & $-$30{,}404 & $-$32{,}804 & $-$27{,}166 & $-$34{,}561 & $-$32{,}453 \\
 & & 1.00 & $-$29{,}670 & $-$27{,}953 & $-$29{,}843 & $-$32{,}515 & $-$27{,}858 & $-$31{,}098 & $-$38{,}929 \\
\cmidrule(l){2-10}
 & \multirow{4}{*}{100}
 & 0.25 & $-$93{,}589 & $-$121{,}757 & $-$121{,}738 & $-$115{,}853 & $-$125{,}177 & $-$137{,}432 & $-$172{,}032 \\
 & & 0.50 & $-$85{,}877 & $-$109{,}364 & $-$95{,}132 & $-$97{,}828 & $-$91{,}889 & $-$127{,}456 & $-$160{,}446 \\
 & & 0.75 & $-$74{,}247 & $-$73{,}888 & $-$84{,}070 & $-$97{,}328 & $-$98{,}135 & $-$126{,}014 & $-$166{,}652 \\
 & & 1.00 & $-$66{,}000 & $-$71{,}780 & $-$81{,}266 & $-$97{,}688 & $-$99{,}058 & $-$124{,}094 & $-$159{,}849 \\
\cmidrule(l){2-10}
 & \multirow{4}{*}{200}
 & 0.25 & $-$188{,}834 & $-$252{,}462 & $-$252{,}596 & $-$249{,}113 & $-$271{,}462 & $-$308{,}710 & $-$395{,}360 \\
 & & 0.50 & $-$173{,}572 & $-$228{,}061 & $-$204{,}752 & $-$215{,}970 & $-$223{,}022 & $-$301{,}318 & $-$374{,}650 \\
 & & 0.75 & $-$151{,}407 & $-$157{,}952 & $-$185{,}759 & $-$216{,}038 & $-$221{,}887 & $-$288{,}418 & $-$374{,}522 \\
 & & 1.00 & $-$134{,}992 & $-$152{,}518 & $-$171{,}996 & $-$218{,}244 & $-$223{,}822 & $-$293{,}808 & $-$382{,}171 \\
\midrule
\multirow{20}{*}{$W$}
 & \multirow{4}{*}{$0.25\times$}
 & 0.25 & $-$47{,}001 & $-$60{,}190 & $-$61{,}343 & $-$59{,}328 & $-$62{,}183 & $-$74{,}384 & $-$84{,}667 \\
 & & 0.50 & $-$42{,}689 & $-$54{,}195 & $-$47{,}726 & $-$49{,}755 & $-$48{,}590 & $-$64{,}705 & $-$82{,}633 \\
 & & 0.75 & $-$36{,}987 & $-$36{,}934 & $-$42{,}920 & $-$50{,}475 & $-$49{,}889 & $-$63{,}953 & $-$83{,}310 \\
 & & 1.00 & $-$32{,}990 & $-$36{,}101 & $-$40{,}457 & $-$49{,}208 & $-$48{,}202 & $-$64{,}134 & $-$82{,}931 \\
\cmidrule(l){2-10}
 & \multirow{4}{*}{$0.5\times$}
 & 0.25 & $-$46{,}449 & $-$58{,}308 & $-$57{,}732 & $-$53{,}548 & $-$57{,}368 & $-$58{,}972 & $-$71{,}611 \\
 & & 0.50 & $-$42{,}297 & $-$52{,}491 & $-$43{,}910 & $-$44{,}451 & $-$41{,}069 & $-$51{,}838 & $-$64{,}612 \\
 & & 0.75 & $-$36{,}194 & $-$34{,}388 & $-$39{,}180 & $-$44{,}281 & $-$39{,}287 & $-$50{,}763 & $-$64{,}222 \\
 & & 1.00 & $-$32{,}305 & $-$33{,}178 & $-$37{,}594 & $-$43{,}718 & $-$41{,}891 & $-$52{,}790 & $-$65{,}101 \\
\cmidrule(l){2-10}
 & \multirow{4}{*}{$1\times$}
 & 0.25 & $-$45{,}243 & $-$54{,}905 & $-$52{,}774 & $-$43{,}626 & $-$45{,}767 & $-$42{,}880 & $-$44{,}720 \\
 & & 0.50 & $-$41{,}253 & $-$48{,}705 & $-$36{,}715 & $-$32{,}699 & $-$27{,}221 & $-$33{,}963 & $-$36{,}374 \\
 & & 0.75 & $-$34{,}623 & $-$29{,}439 & $-$30{,}404 & $-$32{,}804 & $-$27{,}166 & $-$34{,}422 & $-$32{,}453 \\
 & & 1.00 & $-$29{,}670 & $-$27{,}953 & $-$29{,}843 & $-$32{,}515 & $-$27{,}858 & $-$31{,}554 & $-$38{,}929 \\
\cmidrule(l){2-10}
 & \multirow{4}{*}{$2\times$}
 & 0.25 & $-$42{,}939 & $-$50{,}978 & $-$45{,}293 & $-$35{,}160 & $-$36{,}816 & $-$32{,}481 & $-$32{,}178 \\
 & & 0.50 & $-$39{,}015 & $-$44{,}932 & $-$30{,}233 & $-$26{,}649 & $-$23{,}428 & $-$24{,}702 & $-$24{,}005 \\
 & & 0.75 & $-$32{,}506 & $-$25{,}730 & $-$26{,}004 & $-$25{,}547 & $-$23{,}345 & $-$26{,}330 & $-$21{,}497 \\
 & & 1.00 & $-$27{,}765 & $-$24{,}028 & $-$22{,}466 & $-$25{,}884 & $-$23{,}747 & $-$26{,}274 & $-$28{,}859 \\
\cmidrule(l){2-10}
 & \multirow{4}{*}{$4\times$}
 & 0.25 & \textbf{$-$40{,}793} & \textbf{$-$48{,}671} & \textbf{$-$41{,}806} & \textbf{$-$32{,}274} & \textbf{$-$33{,}334} & \textbf{$-$28{,}786} & \textbf{$-$26{,}254} \\
 & & 0.50 & \textbf{$-$36{,}968} & \textbf{$-$42{,}788} & \textbf{$-$28{,}842} & \textbf{$-$23{,}252} & \textbf{$-$19{,}243} & \textbf{$-$21{,}624} & \textbf{$-$17{,}347} \\
 & & 0.75 & \textbf{$-$30{,}975} & \textbf{$-$24{,}601} & \textbf{$-$24{,}525} & \textbf{$-$23{,}981} & \textbf{$-$21{,}680} & \textbf{$-$21{,}052} & \textbf{$-$20{,}977} \\
 & & 1.00 & \textbf{$-$26{,}526} & \textbf{$-$23{,}145} & \textbf{$-$21{,}816} & \textbf{$-$23{,}326} & \textbf{$-$21{,}246} & \textbf{$-$22{,}033} & \textbf{$-$17{,}648} \\
\bottomrule\end{tabular}}
\setlength{\tabcolsep}{6pt}
\end{table}

\begin{table}[htbp]\centering
\setlength{\tabcolsep}{4pt}
\renewcommand{\arraystretch}{1.1}
\caption{One-at-a-time sensitivity analysis on the \texttt{ttd300} benchmark, Part~2 ($v_{\min}$ and $\phi$), continuing from Table~\ref{tab:ttd-oat-a}. As in Part~1, parameter blocks evaluate variations from the shared baseline ($R = 50$, $W = 1\times$, $v_{\min} = 0.1$, $\phi = 2$). Within each value, the four rows represent the endurance fractions $f \in \{0.25, 0.5, 0.75, 1.0\}$, and each entry is the objective $G$ derived from a single run on the first layout (L1). The highest $G$ in each (block, size, $f$) subset is in bold.}
\label{tab:ttd-oat-b}\vspace{4pt}
\scriptsize
\resizebox{\tabfit}{!}{%
\begin{tabular}{@{}l l c r r r r r r r@{}}
\toprule
Param & Value & $f$ & $N=10$ & $N=20$ & $N=30$ & $N=40$ & $N=50$ & $N=75$ & $N=100$ \\
\midrule
\multirow{20}{*}{$v_{\min}$}
 & \multirow{4}{*}{0.05}
 & 0.25 & $-$45{,}335 & $-$55{,}161 & $-$54{,}348 & $-$45{,}032 & $-$46{,}223 & $-$45{,}123 & $-$46{,}756 \\
 & & 0.50 & $-$41{,}376 & $-$48{,}976 & $-$36{,}955 & $-$34{,}944 & $-$29{,}299 & $-$35{,}728 & $-$38{,}455 \\
 & & 0.75 & $-$34{,}792 & $-$29{,}434 & $-$32{,}166 & $-$33{,}620 & $-$29{,}972 & $-$33{,}541 & $-$40{,}192 \\
 & & 1.00 & $-$29{,}826 & $-$28{,}415 & $-$30{,}050 & $-$33{,}627 & $-$29{,}838 & $-$36{,}160 & $-$41{,}311 \\
\cmidrule(l){2-10}
 & \multirow{4}{*}{0.1}
 & 0.25 & $-$45{,}243 & $-$54{,}905 & $-$52{,}774 & $-$43{,}626 & $-$45{,}767 & $-$43{,}424 & $-$44{,}609 \\
 & & 0.50 & $-$41{,}253 & $-$48{,}705 & $-$36{,}715 & $-$32{,}699 & $-$27{,}221 & $-$33{,}963 & $-$37{,}082 \\
 & & 0.75 & $-$34{,}623 & $-$29{,}439 & $-$30{,}404 & $-$32{,}804 & $-$27{,}166 & $-$34{,}561 & $-$32{,}101 \\
 & & 1.00 & $-$29{,}670 & $-$27{,}953 & $-$29{,}843 & $-$32{,}515 & $-$27{,}858 & $-$30{,}452 & $-$38{,}929 \\
\cmidrule(l){2-10}
 & \multirow{4}{*}{0.25}
 & 0.25 & $-$44{,}742 & $-$54{,}042 & $-$50{,}024 & $-$40{,}760 & $-$42{,}799 & $-$39{,}960 & $-$38{,}054 \\
 & & 0.50 & $-$40{,}806 & $-$47{,}825 & $-$34{,}445 & $-$30{,}545 & $-$27{,}692 & $-$31{,}188 & $-$31{,}741 \\
 & & 0.75 & $-$34{,}115 & $-$28{,}361 & $-$27{,}580 & $-$30{,}803 & $-$27{,}138 & $-$29{,}787 & $-$35{,}494 \\
 & & 1.00 & $-$29{,}193 & $-$26{,}566 & $-$27{,}836 & $-$31{,}814 & $-$27{,}747 & $-$32{,}150 & $-$33{,}469 \\
\cmidrule(l){2-10}
 & \multirow{4}{*}{0.5}
 & 0.25 & $-$43{,}275 & $-$51{,}528 & $-$46{,}134 & $-$36{,}177 & $-$37{,}648 & $-$35{,}173 & $-$30{,}297 \\
 & & 0.50 & $-$39{,}388 & $-$45{,}440 & $-$32{,}142 & $-$27{,}347 & $-$21{,}877 & $-$25{,}930 & $-$24{,}447 \\
 & & 0.75 & $-$32{,}808 & $-$26{,}669 & $-$26{,}650 & $-$27{,}184 & $-$24{,}539 & $-$25{,}994 & $-$25{,}895 \\
 & & 1.00 & $-$28{,}068 & $-$25{,}097 & $-$24{,}665 & $-$26{,}889 & $-$25{,}576 & $-$24{,}796 & $-$27{,}964 \\
\cmidrule(l){2-10}
 & \multirow{4}{*}{0.9}
 & 0.25 & \textbf{$-$39{,}611} & \textbf{$-$47{,}495} & \textbf{$-$40{,}244} & \textbf{$-$31{,}295} & \textbf{$-$31{,}237} & \textbf{$-$25{,}226} & \textbf{$-$21{,}388} \\
 & & 0.50 & \textbf{$-$35{,}886} & \textbf{$-$41{,}636} & \textbf{$-$27{,}827} & \textbf{$-$22{,}264} & \textbf{$-$19{,}944} & \textbf{$-$20{,}600} & \textbf{$-$18{,}194} \\
 & & 0.75 & \textbf{$-$30{,}206} & \textbf{$-$24{,}118} & \textbf{$-$23{,}022} & \textbf{$-$22{,}090} & \textbf{$-$20{,}008} & \textbf{$-$19{,}204} & \textbf{$-$17{,}421} \\
 & & 1.00 & \textbf{$-$25{,}912} & \textbf{$-$22{,}308} & \textbf{$-$21{,}010} & \textbf{$-$22{,}697} & \textbf{$-$20{,}705} & \textbf{$-$19{,}533} & \textbf{$-$18{,}705} \\
\midrule
\multirow{20}{*}{$\phi$}
 & \multirow{4}{*}{0.5}
 & 0.25 & \textbf{$-$45{,}243} & $-$60{,}872 & $-$59{,}666 & $-$55{,}520 & $-$56{,}178 & $-$58{,}583 & $-$65{,}018 \\
 & & 0.50 & $-$43{,}646 & $-$57{,}580 & $-$55{,}512 & $-$52{,}150 & $-$54{,}751 & $-$56{,}289 & $-$62{,}420 \\
 & & 0.75 & $-$41{,}515 & $-$50{,}336 & $-$54{,}492 & $-$50{,}573 & $-$53{,}983 & $-$54{,}266 & $-$61{,}686 \\
 & & 1.00 & $-$40{,}142 & $-$48{,}908 & $-$51{,}687 & $-$51{,}382 & $-$53{,}072 & $-$55{,}126 & $-$57{,}656 \\
\cmidrule(l){2-10}
 & \multirow{4}{*}{1}
 & 0.25 & \textbf{$-$45{,}243} & $-$58{,}263 & $-$57{,}224 & $-$49{,}771 & $-$50{,}965 & $-$50{,}648 & $-$56{,}310 \\
 & & 0.50 & $-$42{,}123 & $-$52{,}122 & $-$47{,}023 & $-$42{,}051 & $-$40{,}610 & $-$46{,}723 & $-$51{,}966 \\
 & & 0.75 & $-$38{,}215 & $-$47{,}050 & $-$44{,}406 & $-$41{,}720 & $-$41{,}499 & $-$45{,}283 & $-$51{,}362 \\
 & & 1.00 & $-$36{,}644 & $-$41{,}574 & $-$40{,}681 & $-$40{,}532 & $-$41{,}915 & $-$46{,}100 & $-$50{,}634 \\
\cmidrule(l){2-10}
 & \multirow{4}{*}{2}
 & 0.25 & \textbf{$-$45{,}243} & $-$54{,}905 & $-$52{,}774 & $-$43{,}626 & $-$45{,}767 & $-$43{,}424 & $-$44{,}253 \\
 & & 0.50 & \textbf{$-$41{,}253} & $-$48{,}705 & $-$36{,}715 & $-$32{,}699 & $-$27{,}173 & $-$34{,}460 & $-$38{,}322 \\
 & & 0.75 & $-$34{,}623 & $-$29{,}439 & $-$30{,}404 & $-$32{,}804 & $-$27{,}166 & $-$34{,}539 & $-$32{,}453 \\
 & & 1.00 & $-$29{,}670 & $-$27{,}953 & $-$29{,}843 & $-$32{,}515 & $-$27{,}858 & $-$31{,}554 & $-$38{,}740 \\
\cmidrule(l){2-10}
 & \multirow{4}{*}{3}
 & 0.25 & \textbf{$-$45{,}243} & \textbf{$-$54{,}738} & $-$51{,}843 & $-$42{,}679 & \textbf{$-$42{,}851} & \textbf{$-$41{,}491} & $-$40{,}413 \\
 & & 0.50 & \textbf{$-$41{,}253} & $-$47{,}410 & $-$33{,}952 & \textbf{$-$30{,}772} & $-$27{,}115 & $-$26{,}988 & $-$27{,}590 \\
 & & 0.75 & \textbf{$-$33{,}877} & $-$24{,}857 & $-$24{,}689 & $-$28{,}264 & $-$24{,}488 & $-$27{,}906 & $-$35{,}454 \\
 & & 1.00 & $-$28{,}486 & $-$22{,}423 & $-$23{,}336 & $-$27{,}742 & $-$21{,}902 & $-$27{,}111 & $-$34{,}006 \\
\cmidrule(l){2-10}
 & \multirow{4}{*}{4}
 & 0.25 & \textbf{$-$45{,}243} & \textbf{$-$54{,}738} & \textbf{$-$51{,}352} & \textbf{$-$42{,}389} & $-$43{,}657 & $-$42{,}454 & \textbf{$-$37{,}918} \\
 & & 0.50 & \textbf{$-$41{,}253} & \textbf{$-$45{,}270} & \textbf{$-$32{,}381} & $-$31{,}311 & \textbf{$-$21{,}156} & \textbf{$-$22{,}188} & \textbf{$-$25{,}723} \\
 & & 0.75 & \textbf{$-$33{,}877} & \textbf{$-$22{,}124} & \textbf{$-$23{,}682} & \textbf{$-$26{,}545} & \textbf{$-$19{,}243} & \textbf{$-$23{,}120} & \textbf{$-$25{,}516} \\
 & & 1.00 & \textbf{$-$28{,}376} & \textbf{$-$18{,}461} & \textbf{$-$21{,}826} & \textbf{$-$25{,}136} & \textbf{$-$19{,}529} & \textbf{$-$22{,}398} & \textbf{$-$23{,}416} \\
\bottomrule\end{tabular}}
\setlength{\tabcolsep}{6pt}
\end{table}

\newpage
\section*{S6. The endurance axis on ttd300 instances}

Section~6.2.2 of the main paper evaluates the endurance axis using the reference solutions themselves, as no exact method can certify optimality on the \texttt{ttd300} benchmark. Table~\ref{tab:ttd300-endurance} presents these reference values in full. For each instance size $N$ and endurance fraction $f$, it details the net objective $G$, the makespan $\tau_{\depotend}$ which dictates the rental cost, and the point in the computational budget $B(N)$ at which the search last improved. Each cell represents an average across the five layouts: $G$ and $\tau_{\depotend}$ are derived from the reference solution of each layout, and the time to the best solution is averaged over every run for that configuration. Figure~5 of the main paper plots the data from the first and third blocks.

\begin{table}[htbp]
\centering
\caption{Full-budget SA reference solutions for the \texttt{ttd300} instances, categorised by size $N$ and endurance fraction $f$. The table reports the net objective $G$, the makespan $\tau_{\depotend}$, and the time at which the search last improved (expressed as a percentage of the computational budget $B(N)$).}
\label{tab:ttd300-endurance}
\renewcommand{\arraystretch}{1.15}
\setlength{\tabcolsep}{4pt}
\small
\begin{tabular}{@{}lrrrrrrr@{}}
\toprule
$f$ & $N=10$ & $N=20$ & $N=30$ & $N=40$ & $N=50$ & $N=75$ & $N=100$ \\
\midrule
\multicolumn{8}{@{}l}{\emph{Net objective $G$}} \\
0.25 & $-$41{,}934 & $-$51{,}283 & $-$47{,}430 & $-$48{,}082 & $-$47{,}842 & $-$36{,}661 & $-$35{,}002 \\
0.50 & $-$38{,}444 & $-$39{,}297 & $-$33{,}850 & $-$33{,}602 & $-$31{,}221 & $-$29{,}438 & $-$29{,}890 \\
0.75 & $-$31{,}818 & $-$28{,}656 & $-$29{,}228 & $-$32{,}053 & $-$30{,}681 & $-$30{,}538 & $-$29{,}639 \\
1.00 & $-$25{,}790 & $-$27{,}292 & $-$29{,}213 & $-$31{,}902 & $-$31{,}671 & $-$30{,}654 & $-$31{,}986 \\
\addlinespace
\multicolumn{8}{@{}l}{\emph{Makespan $\tau_{\depotend}$}} \\
0.25 & 917 & 1{,}267 & 1{,}372 & 1{,}595 & 1{,}740 & 1{,}980 & 2{,}388 \\
0.50 & 853 & 1{,}060 & 1{,}155 & 1{,}335 & 1{,}445 & 1{,}838 & 2{,}292 \\
0.75 & 747 & 901 & 1{,}071 & 1{,}314 & 1{,}437 & 1{,}860 & 2{,}285 \\
1.00 & 651 & 867 & 1{,}064 & 1{,}305 & 1{,}449 & 1{,}868 & 2{,}334 \\
\addlinespace
\multicolumn{8}{@{}l}{\emph{Time to best solution (\% of $B(N)$)}} \\
0.25 & 0 & 5 & 62 & 62 & 74 & 92 & 100 \\
0.50 & 0 & 36 & 64 & 58 & 85 & 88 & 93 \\
0.75 & 0 & 89 & 97 & 70 & 76 & 92 & 99 \\
1.00 & 5 & 95 & 94 & 65 & 80 & 100 & 98 \\
\bottomrule
\end{tabular}
\end{table}

\end{document}